%% file: main.tex
\pdfoutput=1
\documentclass[11pt]{article}

\usepackage[preprint]{acl}

\usepackage{times}
\usepackage{latexsym}

\usepackage[T1]{fontenc}
\usepackage[utf8]{inputenc}

\usepackage{microtype}

\usepackage{inconsolata}

\usepackage{graphicx}

\usepackage{minted}
\setminted{
  breaklines=true,
  breaksymbol={},
  breaksymbolleft={},
  breaksymbolright={},
  linenos=false,
  bgcolor=gray!10,
  frame=leftline,
  framerule=0.5pt,
  framesep=3mm,
}
\usepackage{etoolbox}
\BeforeBeginEnvironment{minted}{\vspace{10em}}
\usepackage{tocloft}

\usepackage{amssymb}
\usepackage{subcaption}
\usepackage{booktabs}
\usepackage{enumitem}
\usepackage{xspace}
\usepackage[dvipsnames]{xcolor}
\usepackage{pifont}
\usepackage{tabularx}
\usepackage{colortbl}
\definecolor{tableheader}{HTML}{EFEFEF}
\usepackage{multirow}
\usepackage{listings}
\usepackage{amsmath}
\usepackage{hyperref}

\newcommand{\dset}{SemEval-STM\xspace}

\newcommand{\dbta}{\textsc{DBTA}\xspace}
\newcommand{\qbta}{\textsc{SBTA}\xspace}

\newcommand{\ie}{\textit{i.e.}}
\newcommand{\eg}{\textit{e.g.}}

\definecolor{illiniorange}{HTML}{FF5F05}
\NewDocumentCommand{\tk}
{ mO{} }{\textcolor{illiniorange}{\textsuperscript{\textit{TK}}\textsf{\small{#1}}}}

\addtocontents{toc}{\protect\setcounter{tocdepth}{-10}}

\makeatletter
\renewcommand{\@fnsymbol}[1]{}
\makeatother

\title{{From Documents to Segments: \\A Contextual Reformulation for Topic Assignment}}

\author{
Hoonsang Yoon$^{1,\ast}$\footnotemark[1] \qquad Takyoung Kim$^{2,\ast}$ \qquad Wonkee Lee$^1$ \qquad Ilmin Cho$^1$ \\ \textbf{Dilek Hakkani-Tür}$^{2,\dagger}$ \qquad \textbf{Stanley Jungkyu Choi}$^{1,\dagger}$ \\ $^1$LG AI Research \quad $^2$University of Illinois Urbana-Champaign \\ \texttt{hoonsang\_yoon@lgresearch.ai} \quad \texttt{tk30@illinois.edu}
}

\begin{document}
\maketitle

\renewcommand{\thefootnote}{$\ast,\dagger\;$}
\footnotetext{Equal contribution.}
\renewcommand{\thefootnote}{\arabic{footnote}}

\input{sections/0_abstract}
\input{sections/1_introduction}

\input{sections/2_relatedwork}
\input{sections/3_method}
\input{sections/4_experiment}

\input{sections/5_conclusion}

% \clearpage
\bibliography{reference}

\clearpage
\input{sections/9_appendix}
\end{document}

%% file: sections/0_abstract.tex
\begin{abstract}
Traditional topic modeling assigns a single topic to each document. In practice, however, many real-world documents, such as product reviews or open-ended survey responses, contain multiple distinct topics. This mismatch often leads to \emph{topic contamination}, where unrelated themes are merged into a single topic, making it difficult to identify documents that truly focus on a specific subject. 
We address this issue by introducing \textbf{segment-based topic allocation (\qbta)}, a reformulation of topic modeling that assigns topics not to entire documents, but to \textbf{segments}: short, coherent spans of text that each express a single theme. By modeling topical structure at the segment level, our approach yields cleaner and more interpretable topics and better supports analysis of multi-theme documents.
To support systematic evaluation, we construct a \textbf{\dset}\footnote{\url{https://huggingface.co/datasets/LG-AI-Research/SemEval-STM}}, a new dataset inspired by aspect-based sentiment analysis. Documents are first decomposed into topical segments using large language models (LLMs), followed by human refinement to ensure segment quality. We also propose a segment-level extension of the word intrusion task, enabling human evaluation of topical coherence at the granularity where topics are actually assigned. 
Across multiple models and evaluation metrics, we show that \qbta improves clustering quality and interpretability. Overall, this work provides a practical, scalable framework for fine-grained topic analysis in heterogeneous text corpora where documents naturally span multiple topics.
\end{abstract}

%% file: sections/1_introduction.tex
\section{Introduction}

Topic modeling is a core technique for discovering latent themes in a text corpus. Classical approaches, such as Latent Dirichlet Allocation (LDA; \citealp{10.5555/944919.944937}), represent each topic as a distribution over words (\eg, \texttt{stock, interest} for an \texttt{economics} topic) and each document as a mixture of these topics (\eg, \texttt{politics} and \texttt{economics} topics within a news article). More recent approaches leverage large language models (LLMs) to generate and validate topic representations, often improving topic quality and interpretability~\citep{stammbach-etal-2023-revisiting, pham-etal-2024-topicgpt}.

\input{figures/overview}

Despite these advances, most topic modeling methods share a fundamental assumption: the \textbf{document} is treated as the basic unit of topical coherence. This assumption conflicts in many real-world settings where documents naturally span multiple, heterogeneous topics. For instance, a single employee opinion survey response may discuss \texttt{compensation}, \texttt{workplace culture}, and \texttt{career growth} within a single paragraph. In such cases, document-level topic models struggle to isolate content related to a specific theme without interference from unrelated topics.

We refer to this limitation as \textit{\textbf{topic contamination}}: inferred topics are diluted by off-topic content because each document typically contains multiple themes. Topic contamination degrades interpretability and undermines practical downstream tasks such as topic summarization or automatic topic labeling~\citep{kozlowski2024generativeaiautomatictopic, wanna2024topictagautomaticannotationnmf}. As illustrated in \autoref{fig:overview} (left), document-based topic allocation (\dbta) retrieves entire documents that contain only a small fraction of the topic of interest, making it difficult to extract focused insights.

To address this problem, we propose \textbf{segment-based topic allocation (\qbta)}, a reformulation of topic modeling that changes the unit of topic assignment. Instead of associating topics with entire documents (\ie, \dbta), \qbta operates over \emph{segments}---short, self-contained spans of text (\eg, sentences or clauses) that each express a coherent idea. As shown in \autoref{fig:overview} (right), this formulation enables the model to retrieve and cluster only the segments relevant to a given topic (\eg, price-related statements), improving topical precision and interpretability. 

Our work is partly inspired by recent LLM-based approaches such as TopicGPT~\citep{pham-etal-2024-topicgpt}, which demonstrate that LLMs can effectively identify topics and supporting evidence segments within documents. However, this treats segments as auxiliary explanations for document-level topics. In contrast, we \textbf{elevate the segment to the primary unit of topic representation and inference}, formally redefining the task of topic modeling itself. This shift from \dbta to \qbta offers multiple advantages:

\begin{itemize}[leftmargin=*]

    \item \textbf{Improved topic purity.} By assigning topics only to semantically focused segments, \qbta filters out unrelated content that would otherwise contaminate document-level topics (\autoref{sec:feasibility}).

    \item \textbf{Fine-grained interpretability.} Segments represent complete semantic units, making topic assignments easier to interpret and inspect than document-level mixtures (\autoref{fig:overview}).

    \item \textbf{Better alignment with practical use cases.} Many real-world workflows require retrieving or analyzing specific statements about a topic (\eg, searching only \texttt{compensation}-related feedback in survey responses), rather than entire multi-topic documents.
\end{itemize}

To evaluate the \qbta approach, we introduce a new dataset, \textbf{\dset} (\autoref{sec:dataset}), inspired by aspect-based sentiment analysis corpora~\citep{pontiki-etal-2016-semeval}. Unlike conventional topic modeling datasets, \dset provides segment-level annotations, enabling fine-grained evaluation of topic assignments. We formalize the \qbta task (\autoref{sec:definition}) and empirically show that segments in \dset exhibit strong topic clustering properties~(\autoref{sec:feasibility}). 
In addition, we extend the standard word intrusion task~\citep{NIPS2009_f92586a2} to the segment level, providing a human-centered evaluation that better captures contextual coherence~(\autoref{sec:qit}). Finally, we benchmark a wide range of topic modeling methods under the \qbta formulation (\autoref{sec:experiment}), offering insights for both future research directions and practical deployment.

%% file: figures/overview.tex
\begin{figure}[t!]
    \centering
    \includegraphics[width=\columnwidth]{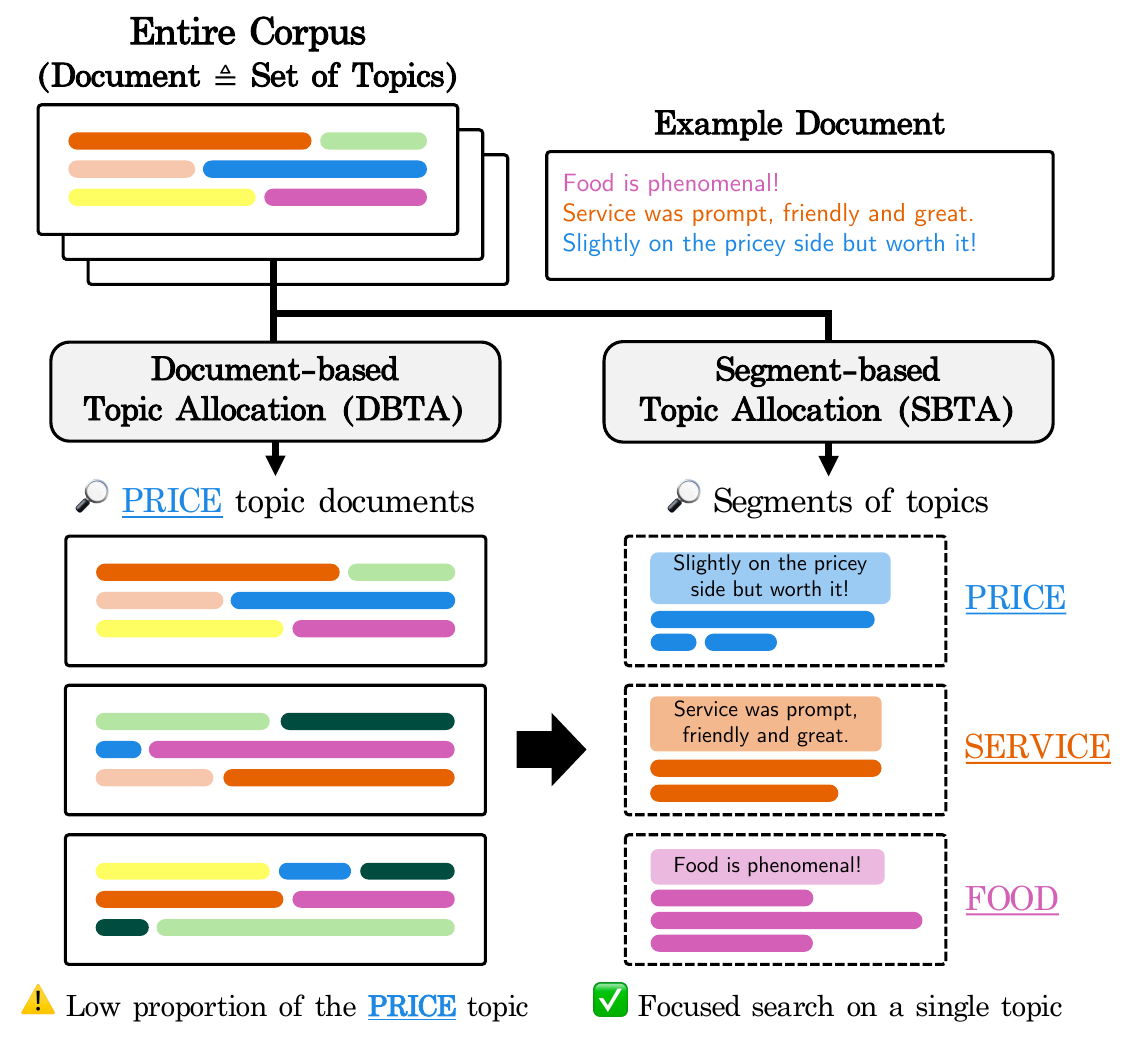}
    \caption{
        Segment-based topic allocation (\qbta) assigns each topic to a specific textual unit (``segment''), thereby improving interpretability and facilitating its effective application across diverse downstream tasks. 
    }
    \label{fig:overview}
    % \vspace{-3.7mm}
\end{figure}

%% file: sections/2_relatedwork.tex
\section{Related Works}

\subsection{Evolution of Topic Modeling Approaches}

Topic modeling aims to uncover latent themes within documents, with LDA~\citep{10.5555/944919.944937} being a seminal method. Although effective, LDA often suffers from interpretability issues and requires manual labeling~\citep{Mei2007AutomaticLO,NIPS2009_f92586a2,Baden-etal-2021-gaps}. Various extensions have been proposed to improve coherence and scalability, including seeded and hierarchical models~\citep{andrzejewski-zhu-2009-latent,Teh01122006}, as well as neural topic models \citep{srivastava2017autoencoding,dieng-etal-2020-topic}. Clustering-based approaches such as BERTopic~\citep{grootendorst2022bertopicneuraltopicmodeling} leverage pre-trained sentence embeddings to construct topic models.

More recently, LLM-based approaches have introduced prompt-driven topic modeling strategies. TopicGPT~\citep{pham-etal-2024-topicgpt} and others~\citep{doi-etal-2024-topic} utilize LLMs to generate and assign topics at the document level, often referring to segment-level content for interpretability. However, in these methods, segments are treated only as explanatory evidence rather than being incorporated as formal units of topic representation.

\subsection{Topic Modeling Evaluation}

\label{sec:evaluation}

Evaluating topic models involves both human and automatic approaches. Early studies introduced word intrusion tasks to assess interpretability~\citep{NIPS2009_f92586a2, newman2010evaluating, mimno-etal-2011-optimizing}. Although these align closely with human judgment, they are costly and difficult to scale.

Metrics such as UCI~\citep{newman2010evaluating}, UMass~\citep{mimno-etal-2011-optimizing}, NPMI~\citep{lau-etal-2014-machine}, and C\textsubscript{v}~\citep{10.1145/2684822.2685324} measure word‐level consistency based on co‐occurrence statistics, later extended with embedding‐based semantic similarity~\citep{10.1145/2911451.2914720, ramrakhiyani-etal-2017-measuring, KORENCIC2018357}. However, most coherence metrics remain word‐centric and often fail to capture document‐level or contextual coherence.

Recent research leverages LLMs for evaluating topic coherence and topic intrusions~\citep{stammbach-etal-2023-revisiting, rahimi-etal-2024-contextualized}. These approaches treat LLMs as proxies for human annotators, achieving greater scalability and interpretability. Yet, most of these methods remain limited to evaluating the quality of topic words alone.

\subsection{Relation with Topic Segmentation}
Topic segmentation identifies points of topical shift within a document to divide it into coherent sections. Early methods such as TextTiling~\citep{10.3115/981732.981734} exploited inter-sentence similarity to detect topic transitions, while TopicTiling~\citep{riedl-biemann-2012-text} extended this approach by incorporating probabilistic topic assignments from LDA to infer segment boundaries. Recent approaches have adopted embeddings and LLMs for improved contextual sensitivity~\citep{fan2024uncoveringpotentialchatgptdiscourse, iacopo_ghinassi_2021_4744399}. Although topic segmentation and our framework both involve textual segments, their objectives are fundamentally different. While topic segmentation focuses on structural partitioning, our segment-based topic modeling treats segments as the analytical unit for deriving cleaner and more faithful topic models, leveraging semantically coherent spans to enhance topic purity, interpretability, and contextual alignment across the corpus.

%% file: sections/3_method.tex
\section{Segment-based Topic Allocation}

\subsection{Task Definition}
\label{sec:definition}

\subsubsection{Basic Notations}

We define the vocabulary as a set of $V$ distinct words, indexed by $\{1,\dots,V\}$. A corpus is denoted by $\mathcal{D}=\{ d_1, \dots, d_D \}$, where each document $d\in\mathcal{D}$ is represented as a sequence of $N_{d}$ word tokens. We assume the existence of $K$ latent topics, indexed by $\{1,\dots,K\}$. 

\subsubsection{Classical Topic Modeling}

In Latent Dirichlet Allocation (LDA; \citealp{10.5555/944919.944937}), the \textbf{topic-word distributions} are represented by probability vectors $\phi_k = (\phi_{k,1},\dots,\phi_{k,V})$ for each topic $k$, where $\phi_k\in\Delta^{V-1}$. Here, $\Delta^{V-1}$ denotes the $(V-1)$-dimensional probability simplex, ensuring that each $\phi_k$ is a valid probability distribution over the vocabulary (\ie, $\phi_{k,v}\ge0$ for all $v$, and $\sum_{v=1}^V{\phi_{k,v}}=1$). The entry $\phi_{k,v}$ corresponds to the probability $P=\{w=v|z=k\}$, where $w\in\{1,\dots,V\}$ denotes a word index from the vocabulary, and $z\in\{1,\dots,K\}$ is a latent topic assignment.

Likewise, the \textbf{document-topic distribution} for each document $d$ is parameterized by $\theta_d=(\theta_{d,1},\dots,\theta_{d,K})\in\Delta^{K-1}$, where $\theta_{d,k}=P(z=k|d)$ indicates the probability of topic $k$ appearing in document $d$. Both $\phi_k$ and $\theta_d$ are drawn from Dirichlet priors, ensuring proper probabilistic constraints.

\subsubsection{Definition of Segment}

Building on the classic topic-word and document-topic formalisms, we introduce a \textbf{topic–segment distribution} that admits \textit{multi-topic} phrases. Unlike traditional topic models such as LDA, which assign a single topic to each word token independently, our model introduces span-level structure by grouping contiguous tokens into segments that may jointly express one or more related topics. 

Formally, a segment in document \(d\) is a pair \(\bigl([i\!:\!j],\,\mathcal{T}\bigr)\) defined as:

\begin{align*}
[i\!:\!j] = \{\,i,\,i+1,\,\dots,\,j\,\}, \;\;
               1 \le i \le j \le N_d,\\ 
\mathcal{T} \subseteq \{1, \dots, K\},\; \mathcal{T} \ne \varnothing
\end{align*}

In other words, a segment is a \textbf{maximal contiguous span of tokens} whose inferred topic labels are entirely contained within a set $\mathcal{T}$. This reflects a key intuition: speakers or writers tend to discuss related topics in compact phrases, such as “\textit{I love the \underline{price} and \underline{quality}},” where $\mathcal{T} = \{\textsc{price}, \textsc{quality}\}$, rather than interleaving many unrelated topics within a short segment. In contrast to prior work such as \citet{arnold-etal-2019-sector}, which segments documents into coarser, multi-sentence units, our segment unit yields more fine-grained topical structure.

\paragraph{Per-document and Corpus-level Collection:}

We gather all segments in document $d$ as:
\begin{align*}
\mathcal{Q}_d = \bigl\{\,Q_d^{(1)}, Q_d^{(2)},\,\dots \bigr\}, \\
Q_d^{(m)} = \bigl([i_m\!:\!j_m],\,\mathcal{T}_m\bigr).
\end{align*}

When a topic-specific view is needed (\eg, if someone is only interested in segments that mention the \texttt{price} topic), we define $\mathcal{Q}_{d,k}= \{\,Q \in \mathcal{Q}_d \mid k\in\mathcal{T}(Q)\,\}$. The \textbf{segment set} for the entire corpus is then given by $\mathcal{Q}= \bigcup_{d\in\mathcal{D}}\mathcal{Q}_d$.

\paragraph{Topical Sparsity:}
Although a document may span a broad mixture of topics, each segment typically focuses on only a small subset. Formally, for a segment $Q=([i\!:\!j],\mathcal{T})$, we usually have $1 \le |\mathcal{T}| \ll |\operatorname{supp}(\boldsymbol{\theta}_d)|$, where $\operatorname{supp}(\boldsymbol{\theta}_d)$ denotes the set of active topics in document $d$. In practice, $|\mathcal{T}|$ rarely exceeds two or three, which aligns with the intuition that most phrases express a tightly coupled semantic focus, rather than blending many disparate themes.

\input{tables/shuffle}

\subsection{Dataset: \dset}
\label{sec:dataset}

To evaluate our reformulated framework, we construct \textbf{\dset}\footnote{Standing for \underline{S}egment-based \underline{T}opic \underline{M}odeling.} dataset by modifying an aspect-based sentiment analysis task that offers a natural testbed for segment-based topic modeling tasks. While conventional topic modeling datasets \citep{hoyle-etal-2022-neural, merity2017regularizingoptimizinglstmlanguage} treat documents as the atomic unit of topical coherence and topics as latent and unannotated, the aspect-based datasets explicitly associate textual spans with predefined \textit{aspects}, interpretable semantic categories such as \texttt{service}, \texttt{food}, \texttt{price} in product reviews. These aspects serve a role analogous to topics in topic modeling, representing interpretable semantic categories that organize document content. Moreover, the aspect labels themselves provide weak supervision that can guide the extraction of segments. By treating aspects as proxy topics, we can prompt an LLM to identify segments in the document that correspond to each aspect. This makes aspect-based datasets a practical and meaningful resource for benchmarking \qbta under semi-supervised or weakly supervised settings. In practice, we employ two domains from the SemEval-2016 ABSA dataset\footnote{Creative Commons Attribution\allowbreak-NonCommercial\allowbreak-ShareAlike 3.0 Unported (CC BY-NC-SA 3.0).}~\citep{pontiki-etal-2016-semeval}: \textit{laptop} and \textit{restaurant}. 

We design two experimental setups, \dbta (conventional formulation) and \qbta (proposed formulation), on top of this dataset to evaluate the feasibility of our segment-based formulation. Adhering to the conventional definition, \dbta setup directly utilizes existing annotations from the SemEval-2016 ABSA dataset, as each document is associated with multiple verified aspects, which serve as topics in our framework. In subsequent sections, we provide a generation and verification pipeline to construct \dset suitable for \qbta.

While several existing benchmarks provide topic annotations at the document level, they are limited in their ability to faithfully evaluate segment-based topic allocation (\qbta). In datasets such as Wiki and Bills \citep{hoyle-etal-2022-neural, merity2017regularizingoptimizinglstmlanguage}, documents often contain a large proportion of content unrelated to the assigned main topic.
In such settings, \qbta may appear trivially advantageous, as isolating a small number of relevant segments from largely off-topic documents is inherently easier.
To enable a fairer comparison, we instead consider a regime where multiple topics are interleaved but topic-irrelevant content does not dominate, allowing the benefits of \qbta to be evaluated more conservatively.
Based on this rationale, we construct a new dataset with explicit segment-level topic annotations.

\subsubsection{Segment Generation}

To instantiate datasets for \qbta in practice, we leverage the generative capabilities of LLMs. For each topic $k \in \{1, \dots, K\}$ and each document $d \in \mathcal{D}$, we query an LLM to identify a set of segments $\mathcal{Q}_{d,k} \subset \mathcal{Q}_d$, where each segment $Q=([i\!:\!j],\mathcal{T})$ satisfies $k\in\mathcal{T}$. That is, we extract maximal contiguous spans within $d$ that are topically coherent and relevant to topic $k$. We use \texttt{o3-mini} in our generation process with the prompt provided in \autoref{sec:promptsegment}.

\subsubsection{Postprocessing and Refinement}

Based on the segment allocation results of the LLM, we discard topics with fewer than 10 segments, reducing the laptop domain from 76 to 33 topics while retaining all 12 restaurant topics, with the same filtering applied to \dbta. Subsequently, the authors manually reassign and merge topics by examining topic-segment pairs; for instance, segments under \texttt{LAPTOP\#GENERAL} are redistributed into more specific topics such as \texttt{LAPTOP\#QUALITY}. This yields a final set of 23 topics for laptop and 11 for restaurant, shared identically across both \dbta and \qbta. \autoref{sec:example_bta} provides postprocessed examples, and \autoref{sec:distribution} illustrates the topic distribution of \dbta and \qbta for each domain.

\subsubsection{Preliminary Experiments}
\label{sec:feasibility}

To validate the feasibility of \dset, we conduct two key experiments: (1) a comparison between SBTA and DBTA on their original topic allocation quality, and (2) a sensitivity test by applying random shuffling to each method’s topic assignments.

We evaluate clustering quality using six standard clustering metrics: \textbf{DB Index}, \textbf{CH Index}, \textbf{MB Score}, \textbf{Silhouette}, \textbf{XB Index}, and \textbf{XB Star}. All metrics are computed using \texttt{all-MiniLM-L6-v2} embeddings\footnote{\url{https://huggingface.co/sentence-transformers/all-MiniLM-L6-v2}. Consistent results with other embeddings are reported in \autoref{sec:shuffle}.}, and for the topic shuffling task, results are averaged over five runs.

\paragraph{(1) Topic allocation quality comparison:}
We first compare the original clustering quality of \qbta and \dbta without any perturbation. As shown in \autoref{tab:shuffle}, \textbf{\qbta achieves higher clustering metric scores than \dbta across all metrics and domains}, suggesting that segment-level topic allocation produces more tightly clustered topic groups when evaluated at the segment granularity. This empirical observation aligns with the core motivation of \qbta: allocating topics to semantically coherent spans (segments) rather than entire documents can yield improved topic purity and structural clarity within this reformulated task framework.

\paragraph{(2) Sensitivity to Topic Shuffling:}
To evaluate how well-structured the original topic assignments are, we introduce a perturbation-based test: topic items (\ie, documents or segments) are randomly shuffled across topics. The underlying assumption is that if the original assignments are semantically meaningful and topologically well-formed, shuffling should significantly degrade clustering performance by disrupting their internal structure.

For example, when documents related to “price” and “design” are originally well-separated, randomly reassigning them to incorrect clusters leads to semantic incoherence (\eg, placing a segment about pricing into the design cluster). Conversely, if the original topic structure is already noisy or loosely defined, such random shuffling introduces minimal additional disorder.

In this setting, the performance degradation from \textbf{SBTA} to the shuffled version, marked with \textbf{SBTA (S)}, is substantially larger than from \textbf{DBTA} to \textbf{DBTA (S)}.\footnote{This can be seen by comparing rows 3–4 against rows 1–2 in Table 1 for each domain.} This demonstrates SBTA’s higher sensitivity due to its inherently more coherent and structured topic clusters. In contrast, DBTA experiences only marginal degradation, sometimes even showing improvements, reflecting a lack of distinct topical structure rather than genuine robustness.

Taken together, these experiments reveal that \textbf{SBTA exhibits stronger clustering metric scores and encodes a finer-grained topic structure compared to DBTA}, one that is more sensitive to disruption when semantic alignment is removed. This suggests that the segment-based reformulation offers distinct advantages for applications requiring fine-grained topical analysis.

\paragraph{On the Limitations of Coherence Metrics:}

As for the coherence metrics (the first four metrics in \autoref{tab:shuffle}), we observe that neither \dbta nor \qbta exhibits consistent performance degradation under topic item shuffling. We attribute this to a fundamental limitation of coherence metrics: they rely on \textit{co-occurrence statistics between word pairs}. In the case of \dbta, the use of full-length documents often leads to frequent words co-occurring broadly across documents, thereby preserving high pairwise co-occurrence scores even when topic items are shuffled. In contrast, for \qbta, the significantly shorter segments reduce the likelihood of strong word co-occurrence patterns being observed in the first place, making it difficult for coherence metrics to distinguish between coherent and incoherent topics, regardless of shuffling. In sum, this behavior of coherence metrics is expected given the underlying characteristics of both \dbta and \qbta.

Hereafter, we utilize the \qbta version of the \dset benchmark in our experiments.

\input{tables/tm_unlabeled}

\subsection{Evaluation: Segment Intrusion Task}
\label{sec:qit}

We additionally introduce a human-centered intrusion evaluation method at the segment level, inspired by traditional word and topic intrusion tasks~\citep{NIPS2009_f92586a2, bhatia-etal-2018-topic}. While previous methods focused on identifying intruder words within topic-word lists, our design extends this approach to the segment level, \textbf{made possible by redefining the unit of topic assignment from documents to segments.}

The core objective of the segment intrusion task is to evaluate {\itshape ``whether a topic has human-identifiable semantic coherence''}~\citep{NIPS2009_f92586a2}. Specifically, either human annotators or LLMs are asked to identify the most semantically divergent segment among a group of candidates, each supposedly sharing the same topical label. A high success rate indicates that the segments are semantically cohesive, validating the quality of topic assignments.

To systematically control task difficulty, we vary both the \textbf{semantic similarity} among candidate segments and \textbf{the number of intruders}. In easy conditions, intruders are drawn from different domains (\eg, spotting a restaurant segment among laptop segments), making them easier to identify. In contrast, hard conditions sample intruders from the same domain, demanding more fine-grained discrimination. The number of intruders is also varied to further modulate task complexity. In this work, we adopt four task variants (Single/Double Intruder with Easy/Hard settings), all newly proposed in this work.

\paragraph{Task 1: Single-Intruder-Easy (SI-E).} 200 sets with five segments from a single domain and one intruder from a different domain. Identify the one intruder segment.

\paragraph{Task 2: Single-Intruder-Hard (SI-H).} 200 sets with five segments and one intruder, all from the same domain. Identify the one intruder segment. \autoref{sec:examplesit} presents an example of this task.

\paragraph{Task 3: Double-Intruder-Easy (DI-E).} 200 sets with four segments from a single domain and two intruders from a different domain. Identify the two intruder segments.

\paragraph{Task 4: Double-Intruder-Hard (DI-H).} 200 sets with four segments and two intruders, all from the same domain. Identify the two intruder segments.

%% file: tables/shuffle.tex
\begin{table*}[t]
\centering \scriptsize
\renewcommand{\arraystretch}{0.8}
\setlength{\tabcolsep}{3pt} 
\caption{
\textbf{Results of the document/segment shuffle test in \dset.} Shuffled results (with mean and standard deviation for 5 repetitions) are marked with \texttt{(S)}. The first four metrics assess coherence based on word frequency, while the remaining metrics evaluate clustering performance based on \textbf{\texttt{all-MiniLM-L6-v2}} embedding. Arrow ($\uparrow$ and $\downarrow$) denote whether higher or lower values indicate better performance, respectively. Metric scores that are negatively affected by the shuffling of documents or segments, consistent with expectations, are highlighted in \textbf{bold}.
}
\begin{tabular*}{\textwidth}{l!{\vrule}cccc!{\vrule}cccccc} 
\toprule
 & NPMI ($\uparrow$) & UMass ($\uparrow$) & UCI ($\uparrow$) & C\textsubscript{v} ($\uparrow$) & DB Index ($\downarrow$) & CH Index ($\uparrow$) & MB Score ($\uparrow$) & Silhouette ($\uparrow$) & XB Index ($\downarrow$) & XB Star ($\downarrow$) \\ \midrule[1pt]

\multicolumn{11}{c}{\textbf{\textit{Domain: Laptop}}} \\ \midrule[1pt]

\dbta & 
-0.0094 & 
-1.4591 & 
-0.3159 & 
0.3984 &
20.1768 &
3.0037 & 
0.0007 & 
-0.0522 & 
95.8645 & 
103.8339 
\\ \midrule

\rowcolor{gray!15} & 
\textbf{-0.0166} &
-1.4129 & 
-0.2576 & 
\textbf{0.3919} &
23.0329 & 
\textbf{0.9943} &
\textbf{0.0002} &
\textbf{-0.0482} &
\textbf{131.2557} &
\textbf{139.1088} \\

\rowcolor{gray!15} \multirow{-2}{*}{\dbta (S)} &
($\pm$ 0.0027) &
($\pm$ 0.0123) &
($\pm$ 0.0314) &
($\pm$ 0.0096) &
($\pm$ 1.5234) &
($\pm$ 0.0225) &
($\pm$ 5.4772e-05) &
($\pm$ 0.0120) &
($\pm$ 17.5548) &
($\pm$ 17.3878) 
\\ \midrule

\qbta &
-0.1626 &
-11.2192 &
-6.9539 &
0.3109 &
6.2767 &
15.5184 &
0.0017 &
0.046 &
10.8348 &
12.4061 
\\ \midrule

\rowcolor{gray!15} &
\textbf{-0.1920} &
-9.5768 &
-5.7639 &
\textbf{0.2862} &
\textbf{28.1171} &
\textbf{1.0172} &
\textbf{0.0003} &
\textbf{-0.0268} &
\textbf{197.1178} &
\textbf{203.2193} \\

\rowcolor{gray!15} \multirow{-2}{*}{\qbta (S)} &
($\pm$ 0.0056) &
($\pm$ 0.1696) &
($\pm$ 0.1844) &
($\pm$ 0.0075) &
($\pm$ 2.1846) &
($\pm$ 0.0607) &
($\pm$ 0.0) &
($\pm$ 0.0045) &
($\pm$ 31.0989) &
($\pm$ 30.3041) 
\\ \midrule[1pt]

\multicolumn{11}{c}{\textbf{\textit{Domain: Restaurant}}} \\ \midrule[1pt]

\dbta &
-0.0034 &
-1.449 &
-0.4289 &
0.3581 &
70.9506 &
1.7204 &
0.001 &
-0.0303 &
1233.5519 &
1264.0482 
\\ \midrule

\rowcolor{gray!15} &
\textbf{-0.0044} &
\textbf{-1.5501} &
-0.2880 &
\textbf{0.3484} &
26.6599 &
\textbf{0.9583} &
\textbf{0.0005} &
-0.0235 &
176.4181 &
180.9349 \\

\rowcolor{gray!15} \multirow{-2}{*}{\dbta (S)} &
($\pm$ 0.0077) &
($\pm$ 0.0805) &
($\pm$ 0.1587) &
($\pm$ 0.0151) &
($\pm$ 1.1914) &
($\pm$ 0.0412) &
($\pm$ 3.7270e-05) &
($\pm$ 0.0025) &
($\pm$ 15.8537) &
($\pm$ 15.0420) 
\\ \midrule

\qbta &
-0.2376 &
-12.2595 &
-8.0276 &
0.3508 &
6.6657 &
22.6709 &
0.005 &
0.0222 &
12.1985 &
13.5955 
\\ \midrule

\rowcolor{gray!15} &
-0.2003 &
-9.6084 &
-5.8915 & 
\textbf{0.2926} &
\textbf{30.8520} &
\textbf{1.0148} &
\textbf{0.0007} &
\textbf{-0.0179} &
\textbf{238.1718} &
\textbf{241.2051} \\

\rowcolor{gray!15} \multirow{-2}{*}{\qbta (S)} &
($\pm$ 0.0261) &
($\pm$ 0.7104) &
($\pm$ 0.6971) &
($\pm$ 0.0171) &
($\pm$ 2.9719) &
($\pm$ 0.0474) &
($\pm$ 5.5902e-05) &
($\pm$ 0.0031) &
($\pm$ 43.5951) &
($\pm$ 44.0460) 
\\ 

\bottomrule[1pt]
\end{tabular*}
\label{tab:shuffle}
% \vspace{-3mm}
\end{table*}

%% file: tables/tm_unlabeled.tex
\begin{table*}[ht]
\centering \scriptsize
\renewcommand{\arraystretch}{0.8}
\setlength{\tabcolsep}{2.8pt} 
\caption{Topic modeling benchmark performance with label-free metrics. Best performance is marked as \textbf{bold}. Full results are demonstrated in \autoref{tab:tm_unlabeled_full}.}
\begin{tabular*}{\textwidth}{l!{\vrule}cccc!{\vrule}cccccc} 
\toprule
 & NPMI ($\uparrow$) & UMass ($\uparrow$) & UCI ($\uparrow$) & C\textsubscript{v} ($\uparrow$) & DB Index ($\downarrow$) & CH Index ($\uparrow$) & MB Score ($\uparrow$) & Silhouette ($\uparrow$) & XB Index ($\downarrow$) & XB Star ($\downarrow$) \\ \midrule[1pt]

\multicolumn{11}{c}{\textbf{\textit{Domain: Laptop}}} \\ \midrule[1pt]

LDA & 
-0.1826 & 
-12.6159 & 
-7.8524 & 
0.3105 &
6.7873 &
4.9365 &
0.0009 & 
-0.0066 & 
10.8622 & 
11.997
\\ \midrule

BERTopic & 
-0.1684 & 
-13.9148 & 
-8.217 &
0.322 &
\textbf{4.5388} &
7.4211 &
0.0034 & 
0.0862 & 
\textbf{4.7237} & 
\textbf{5.3453}

\\ \midrule

\texttt{llama-3.2-3b} & 
\multirow{2}{*}{\textbf{-0.0613}} & 
\multirow{2}{*}{\textbf{-9.4438}} & 
\multirow{2}{*}{\textbf{-4.988}} & 
\multirow{2}{*}{\textbf{0.4024}} &
\multirow{2}{*}{5.8505} & 
\multirow{2}{*}{7.9353} & 
\multirow{2}{*}{\textbf{0.0059}} & 
\multirow{2}{*}{\textbf{0.1647}} & 
\multirow{2}{*}{8.277} & 
\multirow{2}{*}{10.786}
\\ 

\texttt{-instruct-turbo} & 
& & & & & & & &
\\ \midrule

% \texttt{llama-3.3-70b} & 
% \multirow{2}{*}{\underline{-0.1387}} & 
% \multirow{2}{*}{-11.0753} & 
% \multirow{2}{*}{-6.5627} & 
% \multirow{2}{*}{0.2919} &
% \multirow{2}{*}{5.9832} & 
% \multirow{2}{*}{10.047} & 
% \multirow{2}{*}{0.0026} & 
% \multirow{2}{*}{0.0486} & 
% \multirow{2}{*}{10.4182} & 
% \multirow{2}{*}{12.2387} 
% \\ 

% \texttt{-instruct-turbo} & 
% & & & & & & & &
% \\ \midrule

% \texttt{llama-4-maverick-17b} & 
% \multirow{2}{*}{-0.1617} & 
% \multirow{2}{*}{-11.4891} & 
% \multirow{2}{*}{-6.9562} & 
% \multirow{2}{*}{0.3012} & 
% \multirow{2}{*}{6.3082} & 
% \multirow{2}{*}{9.7712} & 
% \multirow{2}{*}{0.0018} & 
% \multirow{2}{*}{0.0405} & 
% \multirow{2}{*}{11.0169} & 
% \multirow{2}{*}{13.4163} 
% \\ 

% \texttt{-128e-instruct-fp8} & 
% & & & & & & & &
% \\ \midrule

\texttt{deepseek-v3} & 
-0.1505 & 
-11.2648 & 
-6.855 & 
0.3069 &
6.9381 &
\textbf{10.7089} & 
0.002 & 
0.0428 & 
10.5334 & 
12.8404
\\ \midrule[1pt]

\multicolumn{11}{c}{\textbf{\textit{Domain: Restaurant}}} \\ \midrule[1pt]

LDA & 
-0.2512 & 
-13.9977 & 
-9.2477 & 
0.3385 &
7.7317 & 
3.3516 & 
0.002 & 
-0.0072 & 
14.8073 &
15.9644
\\ \midrule

BERTopic & 
\textbf{-0.1622} & 
-13.6309 & 
-8.1259 & 
0.3107 &
\textbf{3.9735} & 
6.7305 & 
0.0116 & 
0.1027 & 
\textbf{3.614} &
\textbf{4.0662}
\\ \midrule

% \texttt{4.1-nano} & 
% - & 
% - & 
% - & 
% - &
% - & 
% - & 
% - & 
% - & 
% -
% \\ \midrule

\texttt{gpt-4o} & 
-0.1909 & 
\textbf{-11.0746} & 
-6.9674 & 
0.3076 &
5.63 &
8.7456 & 
0.0054 & 
0.0221 & 
7.9977 & 
9.1804
\\ \midrule

% \texttt{o3-mini} & 
% -0.2324 & 
% -12.5524 & 
% -8.0825 & 
% 0.332 &
% \underline{4.683} &
% 8.8362 & 
% 0.0057 & 
% 0.0269 & 
% \underline{5.2426} & 
% \underline{6.0082}
% \\ \midrule

\texttt{o4-mini} & 
-0.2271 & 
-12.4864 & 
-8.0232 & 
\textbf{0.3451} &
5.3975 &
8.8302 & 
0.0054 & 
0.0239 & 
7.4767 & 
8.3782
\\ \midrule

% \texttt{claude-3.5} & 
% \multirow{2}{*}{-0.2195} & 
% \multirow{2}{*}{-12.1077} & 
% \multirow{2}{*}{-7.7864} & 
% \multirow{2}{*}{0.2943} &
% \multirow{2}{*}{5.4107} & 
% \multirow{2}{*}{8.8726} & 
% \multirow{2}{*}{0.0071} & 
% \multirow{2}{*}{0.0321} & 
% \multirow{2}{*}{6.7834} & 
% \multirow{2}{*}{7.5376} 
% \\ 

% \texttt{-haiku-20241022} & 
% & & & & & & & &
% \\ \midrule

\texttt{claude-3.7} & 
\multirow{2}{*}{-0.2154} & 
\multirow{2}{*}{-11.7673} & 
\multirow{2}{*}{-7.5305} & 
\multirow{2}{*}{0.3233} & 
\multirow{2}{*}{5.2393} & 
\multirow{2}{*}{\textbf{9.7834}} & 
\multirow{2}{*}{0.0059} & 
\multirow{2}{*}{0.0300} & 
\multirow{2}{*}{5.9620} & 
\multirow{2}{*}{7.1544} 
\\ 

\texttt{-sonnet-20250219} & 
& & & & & & & &
\\ \midrule

% \texttt{gemini-2.0} & 
% \multirow{2}{*}{-0.2160} & 
% \multirow{2}{*}{-12.2063} & 
% \multirow{2}{*}{-7.7869} & 
% \multirow{2}{*}{\underline{0.3444}} & 
% \multirow{2}{*}{7.0630} & 
% \multirow{2}{*}{8.9824} & 
% \multirow{2}{*}{0.0066} & 
% \multirow{2}{*}{0.0277} & 
% \multirow{2}{*}{12.8091} & 
% \multirow{2}{*}{15.0293} 
% \\ 

% \texttt{-flash-lite} & 
% & & & & & & & &
% \\ \midrule

% \texttt{gemini-2.0} & 
% \multirow{2}{*}{-0.2033} & 
% \multirow{2}{*}{-11.925} & 
% \multirow{2}{*}{-7.4767} & 
% \multirow{2}{*}{0.3112} & 
% \multirow{2}{*}{5.9880} & 
% \multirow{2}{*}{9.1912} & 
% \multirow{2}{*}{0.0059} & 
% \multirow{2}{*}{0.0302} & 
% \multirow{2}{*}{9.1393} & 
% \multirow{2}{*}{10.6392} 
% \\ 

% \texttt{-flash} & 
% & & & & & & & &
% \\ \midrule

\texttt{gemini-2.5} & 
\multirow{2}{*}{-0.1933} & 
\multirow{2}{*}{-11.1135} & 
\multirow{2}{*}{\textbf{-6.9446}} & 
\multirow{2}{*}{0.3252} &
\multirow{2}{*}{5.1914} & 
\multirow{2}{*}{9.1775} & 
\multirow{2}{*}{0.0055} & 
\multirow{2}{*}{0.0256} & 
\multirow{2}{*}{6.3861} & 
\multirow{2}{*}{7.3846} 
\\ 

\texttt{-flash-preview-04-17} & 
& & & & & & & &
\\ \midrule

% \texttt{qwen2.5-7b} & 
% \multirow{2}{*}{-0.2049} & 
% \multirow{2}{*}{-11.5481} & 
% \multirow{2}{*}{-8.3208} & 
% \multirow{2}{*}{0.3121} &
% \multirow{2}{*}{5.7103} & 
% \multirow{2}{*}{8.4121} & 
% \multirow{2}{*}{0.0054} & 
% \multirow{2}{*}{0.0208} & 
% \multirow{2}{*}{7.3481} & 
% \multirow{2}{*}{8.5285} 
% \\ 

% \texttt{-instruct-turbo} & 
% & & & & & & & &
% \\ \midrule

% \texttt{qwen2.5-72b} & 
% \multirow{2}{*}{-0.2066} & 
% \multirow{2}{*}{-11.7391} & 
% \multirow{2}{*}{-7.4165} &
% \multirow{2}{*}{0.3107} & 
% \multirow{2}{*}{5.3038} & 
% \multirow{2}{*}{\underline{9.2297}} & 
% \multirow{2}{*}{0.0056} & 
% \multirow{2}{*}{0.0248} & 
% \multirow{2}{*}{6.6172} & 
% \multirow{2}{*}{7.8053} 
% \\ 

% \texttt{-instruct-turbo} & 
% & & & & & & & &
% \\ \midrule

\texttt{llama-3.2-3b} & 
\multirow{2}{*}{-0.1981} & 
\multirow{2}{*}{-11.6759} & 
\multirow{2}{*}{-7.2913} & 
\multirow{2}{*}{0.3152} & 
\multirow{2}{*}{5.9292} & 
\multirow{2}{*}{6.1641} & 
\multirow{2}{*}{\textbf{0.0117}} & 
\multirow{2}{*}{\textbf{0.0339}} & 
\multirow{2}{*}{8.2703} & 
\multirow{2}{*}{9.5635} 
\\ 

\texttt{-instruct-turbo} & 
& & & & & & & &
\\ 

% \texttt{llama-3.3-70b} & 
% \multirow{2}{*}{-0.1976} & 
% \multirow{2}{*}{-11.656} & 
% \multirow{2}{*}{-7.3195} &
% \multirow{2}{*}{0.2904} &
% \multirow{2}{*}{5.2444} & 
% \multirow{2}{*}{7.8648} & 
% \multirow{2}{*}{0.0059} & 
% \multirow{2}{*}{0.0232} & 
% \multirow{2}{*}{6.6544} & 
% \multirow{2}{*}{7.6213}
% \\ 

% \texttt{-instruct-turbo} & 
% & & & & & & & &
% \\ \midrule

% \texttt{llama-4-maverick-17b} & 
% \multirow{2}{*}{-0.2311} & 
% \multirow{2}{*}{-12.313} & 
% \multirow{2}{*}{-7.9958} & 
% \multirow{2}{*}{0.3319} & 
% \multirow{2}{*}{4.9387} & 
% \multirow{2}{*}{8.8113} & 
% \multirow{2}{*}{0.0059} & 
% \multirow{2}{*}{0.025} & 
% \multirow{2}{*}{5.5613} & 
% \multirow{2}{*}{6.3629}
% \\ 

% \texttt{-128e-instruct-fp8} & 
% & & & & & & & &
% \\ \midrule

% \texttt{deepseek-v3} & 
% -0.1941 & 
% -11.4944 & 
% -7.2265 &
% 0.328 &
% 6.0934 &
% 8.8475 & 
% 0.007 & 
% \underline{0.033} & 
% 8.6119 & 
% 10.1566
% \\

\bottomrule[1pt]
\end{tabular*}
\label{tab:tm_unlabeled}
% \vspace{-3mm}
\end{table*}

% \bottomrule[1pt]
% \end{tabular*}
% \caption{Topic modeling benchmark performance with label-free metrics. Best performance is marked as \textbf{bold}, and second-best is \underline{underlined}. Full results are demonstrated in \autoref{tab:tm_unlabeled_full}.}
% \label{tab:tm_unlabeled}
% % \vspace{-3mm}
% \end{table*}

%% file: sections/4_experiment.tex
\input{tables/tm_labeled}

\section{Experiments}
\label{sec:experiment}

\subsection{Benchmarking Task 1: Topic Modeling}
\label{sec:bench1}

To evaluate the \qbta performance on \dset, we present benchmark topic modeling results across diverse approaches. Specifically, we employ LDA~\citep{10.5555/944919.944937}, BERTopic~\citep{grootendorst2022bertopicneuraltopicmodeling}, and LLM-based topic modeling approaches with the prompt demonstrated in \autoref{sec:prompttm}.  We used the MALLET \citep{McCallumMALLET} implementation of LDA with Gibbs sampling and BERTopic with all default hyperparameters, except that the number of topics was set to match that of SBTA. For the LLM-based topic assignment approach, we followed a procedure similar to TopicGPT~\citep{pham-etal-2024-topicgpt}. Specifically, each input segment was paired with a predefined set of candidate label topics, and the model was prompted to select the most relevant topic based on its semantic reasoning. To conduct inference, we utilized a diverse set of models, including GPT, Claude, Gemini, Qwen, Llama, and DeepSeek (see \autoref{sec:implementation_details} for technical details).

While we use a predefined topic list in the \dset for evaluation consistency, real-world deployments may lack such supervision. In those scenarios, the topic modeling process can be extended via two complementary phases: a topic generation phase (\eg, prompting an LLM to traverse the corpus and extract candidate topic labels), as demonstrated in \citet{pham-etal-2024-topicgpt}, and a segment-based assignment phase that allocates each topic not to entire documents but to semantically coherent segments. This design effectively extends the TopicGPT framework into a finer-grained topic model aligned with SBTA.

Regarding evaluation metrics, we utilize all aforementioned label-free metrics, consisting of coherence (\autoref{sec:metric_coherence}) and clustering (\autoref{sec:metric_clustering}) measures. Additionally, since \dset provides topic labels for laptop and restaurant domains, we adopt label-based metrics: Purity, ARI, NMI (\autoref{sec:metric_labeled}), Precision, Recall, and F1. LDA and BERTopic only assign topics to documents without explicit labels, we map each predicted topic to the ground-truth label with the most overlapping documents to enable label-based evaluation.

\subsubsection{Results}

For LLM-based performance, we only report models showing either best or second-best performance in the main text. Full model results are provided in \autoref{sec:tmappendix}.

\paragraph{Label-Free Performance:} \autoref{tab:tm_unlabeled} demonstrates label-free topic modeling performance across multiple models and metrics (full version is demonstrated in \autoref{tab:tm_unlabeled_full}). The results indicate that \textbf{each model exhibits distinct strengths within particular domains, thereby underscoring the heterogeneous challenges posed by \dset}. For instance, smaller models (\eg, BERTopic, \texttt{llama-3.2-3b}) demonstrate superior performance compared to larger models in the laptop domain, whereas GPT-based variants generally achieve strong performance in the restaurant domain. These results suggest that \textbf{smaller models, or specific model families, can be strategically deployed for domain-specific applications.}

\paragraph{Label-Based Performance:} \autoref{tab:tm_labeled} presents the performance of label-based topic modeling approaches (full version is demonstrated in \autoref{tab:tm_labeled_full}). In this setting, larger LLMs, such as \texttt{claude-3.7-sonnet}, \texttt{llama-3.3-70b}, and \texttt{deepseek-v3} exhibit strong performance, whereas traditional methods (\ie, LDA, BERTopic) tend to underperform relative to LLM-based approaches.

\input{figures/qit_sub}

\subsection{Benchmarking Task 2: Segment Intrusion Task}
\label{sec:bench2}

To evaluate the segment intrusion task, we employ the same models used in the topic modeling evaluation (\autoref{sec:bench1}) to identify intruder segment(s) from a set of candidates, serving as a proxy for human judgment. We use the prompt presented in \autoref{sec:promptsi} for SI tasks and \autoref{sec:promptdi} for DI tasks. As an upper bound on performance, we assess human performance by randomly sampling 50 instances (out of 200) for each task and averaging the annotations provided by two human participants~\footnote{The annotations were performed by two of the paper’s authors, who were directly involved in the design of the segment intrusion tasks and were well-acquainted with the annotation criteria. Inter-annotator agreement statistics are reported in \autoref{sec:inter_annotate}.}
 (\ie, 50*4=200 annotations per participant). For evaluation metrics, we report F1 score in the main text, providing Recall and Precision results in \autoref{sec:qitappendix}.

\subsubsection{Results}

\autoref{fig:qitsub} visualizes F1 scores of all segment intrusion tasks in laptop domain, using the best-performing model from each model family. As visualized by the downward trend, the results indicate that both LLMs and human annotators experience greater difficulty with the more challenging tasks (\ie, SI-H and DI-H) compared to the easier ones (\ie, SI-E and DI-E), and that identifying two intruders is generally more difficult than selecting a single intruder. Furthermore, as shown in \autoref{fig:qitfull}, smaller models (\eg, \texttt{llama-3.2-3b}, \texttt{qwen-2.5-7b}, and \texttt{llama-4-maverick-17b}) struggle considerably with segment intrusion tasks. Even among larger and higher-performing models, most still fall short of human-level performance. These findings show the difficulty of the proposed segment intrusion tasks and provide an important future direction to improve model capabilities.

%% file: tables/tm_labeled.tex
\begin{table*}[t]
\centering \small
\renewcommand{\arraystretch}{0.6}
\setlength{\tabcolsep}{12pt} 
\caption{Topic modeling benchmark performance with label-based metrics. Best performance is marked as \textbf{bold}. Full results are demonstrated in \autoref{tab:tm_labeled_full}.}
\begin{tabular*}{\textwidth}{l!{\vrule}ccc!{\vrule}ccc} 
\toprule
 & Precision ($\uparrow$) & Recall ($\uparrow$) & F1 ($\uparrow$) & Purity ($\uparrow$) & ARI ($\uparrow$) & NMI ($\uparrow$) \\ \midrule[1pt]

\multicolumn{7}{c}{\textbf{\textit{Domain: Laptop}}} \\ \midrule[1pt]

LDA & 
0.3602 & 
0.3565 & 
0.3577 & 
0.3770 &
0.1573 & 
0.3344 
\\ \midrule

BERTopic & 
0.5139 & 
0.5084 & 
0.5102 & 
0.5033 &
0.2603 & 
0.5262 
\\ \midrule

\texttt{llama-3.3-70b} & 
\multirow{2}{*}{0.7318} & 
\multirow{2}{*}{0.7213} & 
\multirow{2}{*}{0.7248} & 
\multirow{2}{*}{0.7030} & 
\multirow{2}{*}{\textbf{0.4882}} & 
\multirow{2}{*}{0.6342}
\\ 

\texttt{-instruct-turbo} & 
& & & & & 
\\ \midrule

% \texttt{llama-4-maverick-17b} & 
% \multirow{2}{*}{0.6997} & 
% \multirow{2}{*}{0.6902} & 
% \multirow{2}{*}{0.6934} & 
% \multirow{2}{*}{0.6816} & 
% \multirow{2}{*}{0.4479} & 
% \multirow{2}{*}{0.6112}
% \\ 

% \texttt{-128e-instruct-fp8} & 
% & & & & & 
% \\ \midrule

\texttt{deepseek-v3} & 
\textbf{0.7454} & 
\textbf{0.7347} & 
\textbf{0.7383} & 
\textbf{0.7208} &
0.4814 & 
\textbf{0.6543}
\\ \midrule[1pt]

\multicolumn{7}{c}{\textbf{\textit{Domain: Restaurant}}} \\ \midrule[1pt]

LDA & 
0.4530 & 
0.4505 & 
0.4512 & 
0.4812 &
0.1994 & 
0.2397 
\\ \midrule

BERTopic & 
0.6728 & 
0.6679 & 
0.6692 & 
0.6334 &
0.4593 & 
0.5190 
\\ \midrule

% \texttt{4.1-nano} & 
% - & 
% - & 
% - & 
% - &
% - & 
% - 
% \\ \midrule

% \texttt{gpt-4o} & 
% 0.8045 & 
% 0.8001 & 
% 0.8015 & 
% 0.7932 &
% 0.6377 & 
% 0.6648 
% \\ \midrule

% \texttt{o3-mini} & 
% 0.8276 & 
% 0.8226 & 
% 0.8245 & 
% 0.8068 &
% 0.6648 & 
% 0.6693 
% \\ \midrule

% \texttt{o4-mini} & 
% 0.8172 & 
% 0.8124 & 
% 0.8139 & 
% 0.7984 &
% 0.6384 & 
% 0.6640 
% \\ \midrule

% \texttt{claude-3.5} & 
% \multirow{2}{*}{0.6987} & 
% \multirow{2}{*}{0.6931} & 
% \multirow{2}{*}{0.6948} & 
% \multirow{2}{*}{0.6913} & 
% \multirow{2}{*}{0.5019} & 
% \multirow{2}{*}{0.5888} 
% \\ 

% \texttt{-haiku-20241022} & 
% & & & & & 
% \\ \midrule

\texttt{claude-3.7} & 
\multirow{2}{*}{\textbf{0.8386}} & 
\multirow{2}{*}{\textbf{0.8336}} & 
\multirow{2}{*}{\textbf{0.8353}} & 
\multirow{2}{*}{\textbf{0.8224}} & 
\multirow{2}{*}{\textbf{0.6805}} & 
\multirow{2}{*}{\textbf{0.6943}} 
\\ 

\texttt{-sonnet-20250219} & 
& & & & & 
\\

\bottomrule[1pt]
\end{tabular*}
\label{tab:tm_labeled}
% \vspace{-3mm}
\end{table*}

%% file: figures/qit_sub.tex
\begin{figure}[t!]
    \centering
    \includegraphics[width=\columnwidth]{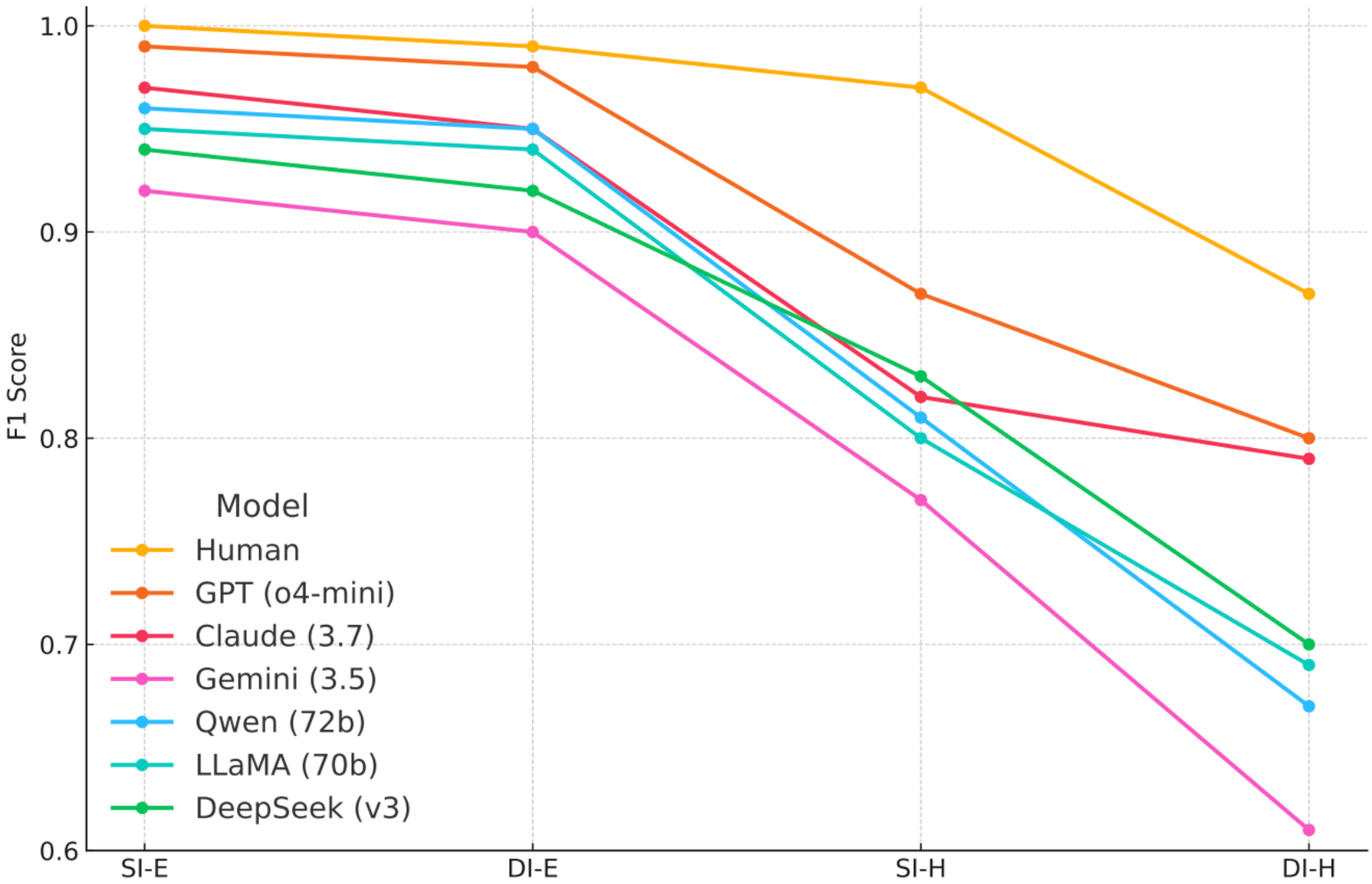}
    \caption{
        Visualized F1 performance of the segment intrusion evaluation in Laptop domain. Human performance is determined by averaging the annotations of two independent participants. Raw performance metrics are reported in \autoref{tab:qit} in \autoref{sec:qitappendix}, and inter-annotator agreement is provided in  \autoref{sec:inter_annotate}.
    }
    \label{fig:qitsub}
    % \vspace{-3.7mm}
\end{figure}

%% file: sections/5_conclusion.tex
\section{Conclusion}

We introduce segment-based topic allocation (\qbta), a shift from document-level topic modeling that assigns topics to the segment level. Through extensive experiments on the \dset dataset, we demonstrate that \qbta substantially improves topic purity and interpretability compared to traditional document-based approaches. Specifically, empirical findings indicate that \qbta yields more distinct topic clusters and exhibits greater sensitivity to underlying topic structure. Additionally, we provide benchmarking results of both label-free and label-based topic modeling evaluations, as well as human-aligned evaluation via the segment intrusion task. Notably, LLMs integrated with \qbta consistently outperform traditional methods across both label-free and label-based metrics, highlighting the synergy between fine-grained topic segmentation and advanced language understanding. Nonetheless, we also observe that certain methods with smaller, more efficient model architectures demonstrate superior performance in domain-specific settings. These results validate \qbta as a practical and scalable solution for fine-grained topic analysis in heterogeneous text corpora.

\section*{Limitations}

While the paper introduces \qbta approach that enhances topic purity and interpretability, several limitations remain. First, the extraction of segments relies heavily on LLMs, introducing potential inconsistencies from automated systems, despite human post-processing. A robust and reliable segment extraction pipeline is required to enable scalable data construction. Furthermore, conventional topic coherence metrics fail to align with \qbta's span-level focus due to reduced word co-occurrence (as analyzed in \autoref{sec:feasibility}), thereby limiting the effectiveness of standard automated evaluations. Lastly, the current dataset primarily contains relatively short documents, reflecting practical scenarios in which users tend to write concise responses (e.g., survey feedback or short reviews). Nevertheless, our data construction pipeline can be extended to corpora with longer documents, following the same segment-based methodology applied in SemEval-STM.

\section*{Ethics Statement}

All models used in our experiments are publicly accessible and are applied in accordance with their intended research use and release terms. Model-specific details are provided in \autoref{sec:implementation_details}.
The proposed SemEval-STM dataset is derived from an existing benchmark through segment-level processing \citep{pontiki-etal-2016-semeval}. This procedure does not introduce new textual content or personally identifiable information. The underlying data consist of short review texts that have been widely used in prior NLP studies and do not involve sensitive or high-risk domains.
AI-based tools were used to assist with writing clarity and editing. All methodological decisions, experiments, and interpretations are carried out and validated by the authors.

%% file: sections/9_appendix.tex
\appendix
\newpage \onecolumn

\addtocontents{toc}{\protect\setcounter{tocdepth}{3}}
\begingroup
\renewcommand{\contentsname}{Appendix Table of Contents}
\setlength{\cftbeforesecskip}{0.5em}
\setlength{\cftbeforesubsecskip}{0.3em}
\setlength{\cftbeforesubsubsecskip}{0.3em}
\hrule
\small
\tableofcontents
\vspace{2em}
\hrule
\vspace{3em}
\endgroup

\clearpage

\section{Details on \dset Construction}

\subsection{Examples of \dbta and \qbta}
\label{sec:example_bta}

\subsubsection{\dbta (Laptop)}
\begin{minted}{json}
{
"CPU#OPERATION_PERFORMANCE": [
    "Other than the slow CPU it works great for everyday use. Heavy gaming is definitely not it's strong point. The cover is a soft rubber texture without the friction. Slick design and looks great.",
    "Being a PC user my whole life.... This computer is absolutely AMAZING!!! 10 plus hours of battery... super fast processor and really nice graphics card.. and plenty of storage with 250 gb(though I will upgrade this and the ram..) This computer is really fast and I'm shocked as to how easy it is to get used to... I've only had mine a day but I'm already used to it... MACS ARE AMAZING!!! GET THIS COMPUTER FOR PORTABILITY AND FAST PROCESSING!!!",
    "not durable . slow processor, just not it",
    "Great laptop from Apple! Apple is unrivaled in terms of build quality and functionality. The retina screen is absolutely beautiful and the touchpad is the best touchpad to date. The pair allow for seamless navigation throughout the whole interface. Battery life is astonishing given the processing power and high resolution display. I can easily get 10 hours out of a full charge. Also, I find OS X much more simple and easy to use than Windows. Overall, a high quality laptop and a great value considering all this computer has to offer.",
    "This computer is absolutely perfect for web browsing, watching video, word processing, and that's about it. If you want a computer that plays games or any advanced tasks like editing video, this computer is NOT for you. Now when I say don't expect to play games on it, I mean it. Pretty much every major game save for Solitare won't even play on it due to the slower processor. Its also worth noting that there no bluetooth, no mircophone jack, no ethernet, no dvd/cd drive/burner, and no usb 3.0 ports. It DOES come with two usb 2.0 ports, a hdmi port, and a webcam. The webcam is nothing special, its there if you need it. Don't expect it to take good pictures or video. The battery is pretty soild. It lasts up to 5-6 hours with normal use. The computer itself runs very quiet and is mostly cool to the touch. The only major negative to me is Windows 8.1 (though YMMV) and the Dell crapware installed. I would strongly recommend to remove it and install Windows 7 on it with a usb drive. The performance will improve on it. Trust me.",
    "This laptop is the most amazing little peice of machinery I have owned outside of the Iphone. It took me a while top get away from the land of PCs, but now that I have, I can't see myself going back to it. The laptop is gorgeous. It is sleek, smooth, and lightweight. It is easily portable and I take it everywhere I go and might require internet access. The screen is very large and crystal clear with amazing colors and resolution. The keyboard is slick and quiet and not bulky like some other laptops I have had in the past. The processor is very quick and effective as I load webpages and applications. It is quiet and a real joy to watch work. I love it and will probably get another one when this goes to the Laptop in the sky!!",
    "Ok, this is probably the best laptop series ever devised by Apple. The case is carved out of a single block of aluminum. Although I opted for the lowest end MacBook Pro, this thing holds its own. The processor screams, and because of the unique way that Apple OSX 16 functions, most of the graphics are routed through the hardware rather than the software. That is how it is able to function better than any other PC. I have recommended this laptop to everyone I know who is buying one. I like it so much, I bought another just for my wife.",
    ...
],
...
}
\end{minted}

\subsubsection{\qbta (Laptop)}
\begin{minted}{json}
{
"POWER_SUPPLY#OPERATION_PERFORMANCE": [
    "HOW DOES THE POWER SUPPLY NOT WORK!!!",
    "the charger stopped working",
    ...
],

"GRAPHICS#DESIGN_FEATURES": [
    "most of the graphics are routed through the hardware rather than the software",
    "it runs anything that doesn't require a dedicated video card",
    ...
],

"LAPTOP#PRICE": [
    "And I'm still paying the bloody financing",
    "All in all, it's well worth its price tag.",
    "For the price and what I get out of it has exceeded my expectations",
    "Good price.",
    "It's not what you pay for.",
    ...
],

"KEYBOARD#DESIGN_FEATURES": [
    "the keyboard is well designed",
    "the lighted keyboard",
    "the user manual explains how to turn on the keyboard  backlight.",
    "I was disappointed when I realized that the keyboard doesn't light up on this model.",
    "a side keyboard. I do a lot of 10-key so I can't be pecking around at the top looking for numbers",
    "the red backlight of the keyboard",
    ...
],
...
}
\end{minted}

\subsubsection{\dbta (Restaurant)}
\begin{minted}{json}
{
"FOOD#QUALITY": [
    "My wife and I ate here earlier this week and have not stopped ranting and raving about the food. If you like spicy food get the chicken vindaloo. It's very spicy but not offensive. We will definitely go back.",
    "The place is a lot of fun. My six year old loved it. The characters really make for an enjoyable experience. The food however, is what one might expect. It is very overpriced and not very tasty. However, I think Jeckll and Hydes t is one of those places that is fun to do once. We had a good time.",
    "One of the BEST Bukhara Grill, the tagline says it all.. \"INDIAN SPICE RAVE\" My GF and I dine at Bukhara often as she lives near it. The lunch buffet is expensive but is deff worth it. We go often for lunch and the place is packed. We have gone for dinner only a few times but the same great quality and service is given. Bukhara is on my top 5 Indian places in NYC",
    "My boyfriend and I went there to celebrate my birthday the other night and all I can say is that it was magnificent. From the spectacular caviar to the hospitable waitstaff, I felt like royalty and enjoyed every second of it. Considering we were the last patrons there and it was after the closing time, the waitstaff did not rush us at all and made us feel comfortable and relaxed. I highly recommend Caviar Russe to anyone who wants delicious top grade caviar and fantastic service.",
    ...
],

"FOOD#STYLE_OPTIONS": [
    "Went here last night - nice decor, good service, but the food was surprisingly excellent. The portions are HUGE, so it might be good to order three things to split (rather than one appetizer and entree per person) for two people. Among all of the new 5th avenue restaurants, this offers by far one of the best values for your money. Can't wait to go back.",
    "They forgot a sandwich, didn't include plastic forks, and didn't include pita with the hummus platter. Also, the sandwiches (nearing $7) didn't come with anything like chips or a side. Overall, not worth the money. Eating in, the atmosphere saves it, but at your desk, it's a very disappointing experience.",
    "I love Indian food and consider myself to be quite an expert on it. Chennai Garden is my favorite Indian restaurant in the city. They have authentic Indian at amazin prices. This restaurant is VEGETARIAN; there are NO MEAT dishes whatsoever. The seats are uncomfortable if you are sitting against the wall on wooden benches. It's a rather cramped and busy restaurant and it closes early.",
    "honestly the worst sushi my husband and i had in our entire lives. believe us, we've been eating sushi for over 15 yrs. not sure why this restaurant would be rated that highly. the all-u-can-eat sushi is definitely in very poor quality. limited menu, no-so-fresh ingredients, thinly-sliced fish, fall-apart rice.  the only things u could really taste are the very salty soy sauce (even its low sodium), the vinegar-soaked rice, and the scallion on top of the fish.  the waitstaffs are nice though. wont come back again for sure!",
    "I recently tried Suan and I thought that it was great. This little place definitely exceeded my expectations and you sure get a lot of food for your money. The service was fast and friendly and the food was very tasty and they had the best hot sauce to add to your meals. I have to say that I am pleasantly suprised and I will most likely stop in again if I am in the neighborhood.",
    "delicious simple food in nice outdoor atmosphere. Kind, attentive wait staff. I really like both the scallops and the mahi mahi (on saffron risotto-yum!). My friend devoured her chicken and mashed potatos. Delicious crab cakes too. Even if the food wasn't this good, the garden is a great place to sit outside and relax. Great neighborhood joint.",
    ...
],
...
}
\end{minted}

\subsubsection{\qbta (Restaurant)}
\begin{minted}{json}
{
"FOOD#QUALITY": [
    "but its the best pie on the UWS!",
    "I'm still mad that i had to pay for lousy food",
    "The food was lousy - too sweet or too salty and the portions tiny.",
    "the food is always consistently, outrageously good",
    "The duck confit is always amazing",
    ...
], 

"RESTAURANT#PRICES": [
    "at mortal prices",
    "without being over-priced.",
    "half off till 8pm",
    "at reasonable prices",
    ...
],

"DRINKS#QUALITY": [
    "go here for the drinks! esp. during happy hour! enough said!",
    "Decent wine at reasonable prices.",
    "a great glass of wine while we waited",
    "the house champagne is a great value",
    "The drinks are always welll made",
    "The sangria was pretty tasty and good on a hot muggy day",
    "wine were excellent",
    "Slightly above average wines",
    ...
],
...
}
\end{minted}

\clearpage

\subsection{Topic Distribution after Postprocessing (Manual Reallocation \& Number Filtering)}
\label{sec:distribution}

\input{figures/distribution}

\clearpage

\section{Quantitative Evaluation Metrics for Topic Modeling}
\label{sec:metrics}

\subsection{Coherence Metrics}
\label{sec:metric_coherence}

Topic coherence metrics aim to quantify the semantic consistency of topic words produced by topic models such as LDA. We used Gensim's CoherenceModel for evaluation with default settings and top-10 words per topic. For LDA, the top-10 words per topic were selected based on the highest probability terms in the topic-word distribution. In the case of BERTopic, we extracted the top-10 words using the class-based TF-IDF (c-TF-IDF) scores, which reflect the importance of words within each topic cluster. For LLM-based topic modeling, the top-10 words were determined by selecting the most frequent terms appearing in the segments assigned to each topic.

Here, we describe three representative coherence metrics used in our experiments.

\subsubsection{NPMI}
Normalized Pointwise Mutual Information (NPMI)~\cite{lau-etal-2014-machine} quantifies the topic coherence based on the co-occurrence statistics between topic words.
Given the top-$N$ words $\{w_1, \dots, w_N\}$ for a topic $k \in \{1, \dots, K\}$, the coherence is computed as the average NPMI over all unique word pairs:
\begin{equation}
\text{Coherence}(k) = \frac{2}{N(N - 1)} \sum_{i=1}^{N - 1} \sum_{j = i + 1}^{N} \text{NPMI}(w_i, w_j).
\label{eq:avg_npmi}
\end{equation}
Each NPMI score $[-1,1]$ quantifies the degree of semantic association between two words, based on their co-occurrence statistics in $\mathcal{D}$:
\begin{equation}
\text{NPMI}(w_i, w_j) = \frac{\log \frac{P(w_i, w_j)}{P(w_i) P(w_j)}}{-\log P(w_i, w_j)},
\label{eq:npmi}
\end{equation}
where $P(w_i)$ and $P(w_i, w_j)$ represent the probabilities of word occurrence and co-occurrence in the corpus $\mathcal{D}$, respectively.

\subsubsection{UMass}
UMass~\citep{mimno-etal-2011-optimizing} is a document-based coherence metric, defined over the range $(-\infty, 0)$---where values closer to 0 indicate higher topic coherence---that measures the co-occurrence frequency of topic word pairs in a reference corpus.
Given the top-$N$ words $\{w_1, \dots, w_N\}$ for a topic $k$, the coherence score is computed as:
\begin{equation}
\mathrm{Coherence}(k) = \sum_{i=2}^{N} \sum_{j=1}^{i-1} \log \frac{D(w_i, w_j) + \epsilon}{D(w_j)},
\label{eq:umass}
\end{equation}
where $D(w_j)$ denotes the number of documents in $\mathcal{D}$ that contain word $w_j$, and $D(w_i, w_j)$ denotes the number of documents in which both $w_i$ and $w_j$ appear; the constant $\epsilon$ is a smoothing term.

\subsubsection{UCI}
UCI~\citep{newman2010evaluating} computes the average pointwise mutual information between all unique pairs of top-$N$ topic words, using co-occurrence counts within a reference corpus. The coherence score is defined as:

\begin{equation}
\mathrm{Coherence}(k) = \frac{2}{N(N - 1)} \sum_{i=1}^{N - 1} \sum_{j=i+1}^{N} \log \frac{P(w_i, w_j)}{P(w_i) P(w_j)}.
\label{eq:uci}
\end{equation}
As in NPMI, $P(w_i)$ and $P(w_i, w_j)$ represent the probabilities of word occurrence and co-occurrence in the corpus $\mathcal{D}$.
Unlike NPMI, which normalizes the score to the range $[-1, 1]$, UCI produces unbounded scores in the range $(-\infty, +\infty)$, where higher values indicate stronger topic coherence.

\subsubsection{\texorpdfstring{C\textsubscript{v}}{Cv}}
C\textsubscript{v}~\citep{10.1145/2684822.2685324} is a composite coherence measure that integrates several desirable aspects from existing metrics, including a sliding window, one-set segmentation, boolean context vectors, and cosine similarity. It is designed to better reflect human interpretability of topic quality by considering both word co-occurrence and contextual similarity.

Given the top-$N$ words $\{w_1, \dots, w_N\}$ for a topic $k$, the coherence score is computed as the average cosine similarity between boolean context vectors for all word pairs:
\begin{equation}
\mathrm{Coherence}(k) = \frac{2}{N(N - 1)} \sum_{i=1}^{N - 1} \sum_{j=i+1}^{N} \cos\left( \vec{v}_{w_i}, \vec{v}_{w_j} \right),
\label{eq:cv}
\end{equation}
where $\vec{v}_{w_i}$ and $\vec{v}_{w_j}$ are the boolean context vectors for words $w_i$ and $w_j$, and $\cos(\cdot)$ denotes the cosine similarity.

Unlike UMass or UCI, C\textsubscript{v} does not depend solely on raw frequency counts but leverages semantic similarity between word contexts. The score ranges from $-\infty$ to $1$, with higher values indicating stronger topic coherence.

\subsection{Clustering Metrics}
\label{sec:metric_clustering}

Clustering-based metrics evaluate topic models that produce topic distributions by measuring how well these distributions form compact and well-separated groups. In this section, we introduce several clustering metrics used in our experiments.

\subsubsection{Dunn Index}
The dunn index~\citep{dunn1974well} evaluates clustering quality based on compactness (intra-cluster distance) and separation (inter-cluster distance). 
A higher dunn score indicates that clusters are both internally tight and well-separated from each other, which is desirable for topic models aiming to produce distinct topics.
Given a set of $k$ topic clusters $\{C_1, C_2, \dots, C_K\}$, the dunn index is defined as:
\begin{equation}
\text{Dunn} = \frac{\displaystyle\min_{1 \leq i < j \leq K} \delta(C_i, C_j)}{\displaystyle\max_{1 \leq k \leq K} \Delta(C_k)},
\label{eq:dunn}
\end{equation}
where $\delta(C_i, C_j)$ indicates the inter-cluster distance, defined as the minimum distance between any two points belonging to different clusters $C_i$ and $C_j$, and $\Delta(C_k)$ denotes the intra-cluster distance, calculated as the maximum distance between any two points within the same cluster $C_k$.

\subsubsection{Davies-Bouldin Index (DB Index)}
The Davies–Bouldin index~\citep{davies2009cluster} quantifies the average similarity between each cluster and its most similar counterpart.
For a set of \( k \) clusters \( \{C_1, C_2, \dots, C_K\} \), the index is defined as:
\begin{equation}
\text{DB} = \frac{1}{K} \sum_{i=1}^{K} \max_{j \ne i} \left( \frac{S_i + S_j}{M_{ij}} \right),
\label{eq:davies}
\end{equation}
where \( S_i \) denotes the intra-cluster distance (e.g., the average distance between points in \( C_i \) and its centroid), and \( M_{ij} \) is the inter-cluster distance between the centroids of clusters \( C_i \) and \( C_j \).
Lower values of the Davies–Bouldin index indicate better clustering, as they reflect low intra-cluster dispersion and high inter-cluster separation.

\subsubsection{Calinski-Harabasz Index (CH Index)}
The Calinski–Harabasz index~\citep{calinski1974dendrite} evaluates clustering quality based on the ratio of inter-cluster dispersion to intra-cluster dispersion; a higher score indicates that clusters are well-separated and internally compact.
Given a clustering of $n$ data points into $k$ clusters, the index is defined as:
\begin{equation}
\text{CH} = \frac{B}{W} \cdot \frac{n - k}{k - 1},
\end{equation}
with
\[
B = \sum_{i=1}^{K} n_i \| c_i - c \|^2, \quad
W = \sum_{i=1}^{K} \sum_{x \in C_i} \| x - c_i \|^2,
\]
where $c_i$ is the centroid of cluster $C_i$, $c$ is the global centroid of $n$ data points, and $n_i$ is the number of data points in $C_i$.
The first term $B / W$ captures the ratio between inter-cluster and intra-cluster dispersion, encouraging large separation and compactness. The second term $(n - k)/(k - 1)$ serves as a scaling factor that penalizes excessive numbers of clusters and helps avoid overfitting.

\subsubsection{Silhouette Score}
This score~\citep{rousseeuw1987silhouettes} measures the extent to which each data point is more similar to its own cluster than to the nearest alternative cluster. Specifically, for $n$ data points, the silhouette score $s$ is defined as:
\begin{equation}
\begin{split}
S &= \frac{1}{n} \sum_{i=1}^n s(i), \\
\text{where} \quad s(i) &= \frac{b(i) - a(i)}{\max(a(i), b(i))}
\end{split}
\label{eq:silhouette}
\end{equation}
where $a(i)$ is the average distance between $x_i$ and all other points in the same cluster (i.e., intra-cluster), $b(i)$ is the minimum average distance between $x_i$ and points in the nearest cluster (i.e., inter-cluster).
The score ranges from $[-1, 1]$; the higher the score, the better the clustering quality.

\subsubsection{Maulik Bandyopadhyay (MB Score)}
The Maulik–Bandyopadhyay index~\citep{maulik2003performance} evaluates clustering quality by jointly considering intra-cluster compactness (i.e., the sum of within-cluster distances) and inter-cluster separation (i.e., the maximum distance between cluster centroids).
It is defined as:
\begin{equation}
\text{MB} = \left( \frac{1}{K} \cdot \frac{E_1}{E_K} \cdot D_K \right)^p,
\label{eq:mb}
\end{equation}
where each component is computed as:
\begin{align}
E_1 &= \sum_{i=1}^{n} \| x_i - c \|^2, \\
E_K &= \sum_{j=1}^{K} \sum_{x_i \in C_j} \| x_i - c_j \|^2, \\
D_K &= \max_{j \ne k} \| c_j - c_k \|.
\end{align}
Here, $n$ is the total number of data points, $K$ is the number of clusters, $C_j$ denotes the $j$-th cluster, $c$ is the global centroid of all $n$ points, and $c_j$ is the centroid of cluster $C_j$. The exponent $p \geq 1$ controls the influence of the separation term, being typically set to $p = 2$.
Higher values indicate better clustering, favoring compact clusters that are well separated.

\subsubsection{Xie Beni Index (XB Index)}
The Xie–Beni index~\citep{xie1991validity} evaluates clustering quality by jointly measuring intra-cluster compactness and inter-cluster separation, being particularly suited to fuzzy clustering algorithms such as fuzzy C-means (FCM), where each data point is assigned a degree of membership $\mu_{ik} \in [0,1]$ to multiple clusters\footnote{In our implementation, the XB index is computed without fuzzy membership weights, i.e., all points are assumed to belong fully to a single cluster. This simplification follows the default behavior of the evaluation tool we used.}.
The index is defined as:
\begin{equation}
\text{XB} = \frac{\sum_{i=1}^{c} \sum_{k=1}^{n} \mu_{ik}^m \| x_k - v_i \|^2}{n \cdot \min_{i \ne j} \|v_i - v_j\|^2},
\label{eq:xb}
\end{equation}
where $v_i$ denotes the centroid of cluster $i$, $\mu_{ik}$ is the degree of membership of point $x_k$ to cluster $i$, $m \geq 1$ is the fuzzifier parameter, and $n$ is the total number of data points. The numerator measures the weighted intra-cluster variance, and the denominator is the minimum squared distance between any pair of cluster centroids; the lower values indicate more compact and better-separated clusters.

\subsubsection{Xie Beni Index Star (XB Star)}
The Xie–Beni Star index\footnote{XB* is a non-standard variant of XB, which is nonetheless implemented in several toolkits, including the one used in our experiments.} modifies the original XB by replacing the average intra-cluster compactness with the worst-case cluster variance, making it more sensitive to poorly formed clusters. Specifically, it replaces the global average intra-cluster variance with the maximum average variance among clusters:
\begin{equation}
\text{XB}^* = \frac{ \max_{1 \leq i \leq K} \left( \frac{1}{|C_i|} \sum_{x_k \in C_i} \| x_k - c_i \|^2 \right) }{ \min_{i \ne j} \| c_i - c_j \|^2 },
\end{equation}
where $C_i$ denotes the $i$-th cluster, $|C_i|$ is its size, $c_i$ is the centroid of $C_i$, and the denominator is the minimum squared distance between any pair of cluster centroids.
This formulation makes XB\* more sensitive to unbalanced clusters, thereby penalizing clustering results that contain even a single weak cluster. As with the original XB index, lower values indicate better clustering.

\subsection{Label-based Metrics}
\label{sec:metric_labeled}

\subsubsection{Purity}

Purity quantifies the extent to which each predicted cluster $\omega_k$ contains data points from a single ground-truth class $c_j$. It is defined as:

\begin{equation}
\text{Purity}(\Omega, \mathbb{C}) = \frac{1}{N} \sum_{k} |\omega_k| \cdot \max_j \frac{|\omega_k \cap c_j|}{|\omega_k|}
\end{equation}

\paragraph{Inverse Purity.}
Also known as \emph{completeness}, Inverse Purity evaluates whether items from the same ground-truth class are grouped together in a single cluster:

\begin{equation}
\text{InversePurity}(\mathbb{C}, \Omega) = \frac{1}{N} \sum_{j} |c_j| \cdot \max_k \frac{|c_j \cap \omega_k|}{|c_j|}
\end{equation}

\paragraph{F1-based $P_1$ Score.}
To balance Purity and Inverse Purity, \citep{10.1007/s10791-008-9066-8} proposed the $P_1$ score, defined as:

\begin{equation}
P_1 = \frac{1}{N} \sum_{k} |c_k| \cdot \max_j F_1(c_j, \omega_k)
\end{equation}

where the $F_1$ score between ground-truth class $c_j$ and cluster $\omega_k$ is:

\begin{equation}
F_1(c_j, \omega_k) = \frac{2 \cdot P(c_j, \omega_k) \cdot R(c_j, \omega_k)}{P(c_j, \omega_k) + R(c_j, \omega_k)}
\end{equation}

with precision and recall defined as:

\begin{equation}
P(c_j, \omega_k) = \frac{|c_j \cap \omega_k|}{|\omega_k|}, \quad R(c_j, \omega_k) = \frac{|c_j \cap \omega_k|}{|c_j|}
\end{equation}

\subsubsection{Adjusted Rand Index}
The Rand Index (RI) measures the agreement between two clusterings by counting pairwise assignments. The Adjusted Rand Index~\citep{RePEc:spr:jclass:v:2:y:1985:i:1:p:193-218, 10.1145/1553374.1553511} corrects the RI for chance agreement:

\begin{equation}
\text{ARI} = \frac{\text{RI} - \mathbb{E}[\text{RI}]}{\max(\text{RI}) - \mathbb{E}[\text{RI}]},
\label{eq:ari}
\end{equation}

where $\mathbb{E}[\text{RI}]$ is the expected Rand Index under a random model. ARI ranges from $-1$ to $1$, with $0$ indicating random assignments and $1$ perfect agreement.

\subsubsection{NMI}

Mutual Information (MI) quantifies the information shared between the cluster assignments $\mathcal{C}$ and ground-truth labels $\mathcal{G}$~\citep{6773024}. The Normalized Mutual Information~\citep{cluster_ensemble} rescales this value:

\begin{equation}
\text{NMI}(\mathcal{C}, \mathcal{G}) = \frac{2 \cdot I(\mathcal{C}; \mathcal{G})}{H(\mathcal{C}) + H(\mathcal{G})},
\label{eq:nmi}
\end{equation}

where $I(\mathcal{C}; \mathcal{G})$ is the mutual information between $\mathcal{C}$ and $\mathcal{G}$, and $H(\cdot)$ denotes the entropy. NMI ranges from $0$ (no shared information) to $1$ (perfect correlation), and is robust to varying numbers of clusters.

\section{Shuffle Test Results with Different Embeddings}
\label{sec:shuffle}

\input{tables/shuffle_appendix}
\clearpage
\section{Prompts}
\label{sec:prompt}

\subsection{Segment Generation for \dset}
\label{sec:promptsegment}

\begin{minted}{markdown}
You are the professional of finding the reason of category allocation.
Within the TEXT, there are quotation or spans that are reason for allocation of CATEGORY.
For each CATEGORY, find the span that can be ground reason for the CATEGORY allocation.
Return the QUOTATION in the format of JSON and only JSON.

[Conditions]
- Multiple QUOTATION can be found with same CATEGORY.
- QUOTATION should be more than two words.
- Try to allocate specific topic as possible rather than GENERAL topic

[Example 1]
TEXT: I get giddy every time I use this thing. It is a thing of beauty and fast enough. What's fast enough? When you click and don't wait. Waiting is horrible, and not waiting is the best thing on earth. So yes, it is ridiculously fast. The battery will get you from LA to NY no problem.
DOMAIN: LAPTOP
CATEGORY: ["LAPTOP#DESIGN_FEATURES", "LAPTOP#OPERATION_PERFORMANCE", "BATTERY#OPERATION_PERFORMANCE"]
QUOTATION: {{"LAPTOP#DESIGN_FEATURES": ["It is a thing of beauty"], "LAPTOP#OPERATION_PERFORMANCE": ["fast enough"], "BATTERY#OPERATION_PERFORMANCE": ["The battery will get you from LA to NY no problem"]}}

[Example 2]
TEXT: This laptop is amazing! Windows 8.1 has its pros and cons. The keyboard is backlit but you have to press the F5 key to turn it on. I was nervous at first that I purchased a lemon but the user manual explains how to turn on the keyboard  backlight. Of course this will decrease your battery life. The sound is very nice and crisp. Also, I have no regrets about going with the 8GB RAM because its super fast. I'm still exploring the various features but overall I am satisfied with my purchase.
DOMAIN: LAPTOP
CATEGORY: [""KEYBOARD#USABILITY", "OS#GENERAL", "MULTIMEDIA_DEVICES#QUALITY", "LAPTOP#GENERAL", "LAPTOP#DESIGN_FEATURES", "LAPTOP#OPERATION_PERFORMANCE", "MEMORY#DESIGN_FEATURES", "KEYBOARD#DESIGN_FEATURES"]
QUOTATION: {{"KEYBOARD#USABILITY": ["The keyboard is backlit but you have to press the F5 key to turn it on"], "OS#GENERAL": ["Windows 8.1 has its pros and cons"], "MULTIMEDIA_DEVICES#QUALITY": ["The sound is very nice and crisp"], "LAPTOP#GENERAL": ["I'm still exploring the various features but overall I am satisfied with my purchase", "This laptop is amazing!"], "LAPTOP#OPERATION_PERFORMANCE": ["Of course this will decrease your battery life"], "MEMORY#QUALITY": ["Also, I have no regrets about going with the 8GB RAM because its super fast"],  "KEYBOARD#DESIGN_FEATURES": ["the user manual explains how to turn on the keyboard backlight"]}}

[Example 3]
TEXT: disadvantages: breakfast is most suitable only for asians, the view, the rooms are rather narrow, the kitchen is open on the lounge.... prices above average.
DOMAIN: HOTEL
CATEGORY: ["FOOD_DRINKS#STYLE_OPTIONS", "ROOMS#DESIGN_FEATURES", "HOTEL#PRICES"]
QUOTATION: {{"FOOD_DRINKS#STYLE_OPTIONS": ["breakfast is most suitable only for asians"], "ROOMS#DESIGN_FEATURES": ["the view, the rooms are rather narrow, the kitchen is open on the lounge"], "HOTEL#PRICES": ["prices above average."]}}

[Example 4]
TEXT: the service was good, the food wasn't bad, the beverage service was really good, except for the plastic cups.
DOMAIN: HOTEL
CATEGORY: ["FOOD_DRINKS#QUALITY", "SERVICE#GENERAL"]
QUOTATION: {{"FOOD_DRINKS#QUALITY": ["the food wasn"t bad"], "SERVICE#GENERAL": ["the service was good", "the beverage service was really good"]}}

[Example 5]
TEXT: we love th pink pony. THe perfect spot. Food-awesome. Service- friendly and attentive. Ambiance- relaxed and stylish. Don't judge this place prima facie, you have to try it to believe it, a home away from home for the literate heart.
DOMAIN: RESTAURANT
CATEGORY: ["RESTAURANT#GENERAL", "FOOD#QUALITY", "SERVICE#GENERAL", "AMBIENCE#GENERAL"]
QUOTATION: {{"RESTAURANT#GENERAL": ["we love th pink pony. THe perfect spot.", "Don't judge this place prima facie, you have to try it to believe it, a home away from home for the literate heart."], "FOOD#QUALITY": ["Food-awesome."], "SERVICE#GENERAL": ["Service- friendly and attentive."], "AMBIENCE#GENERAL": ["Ambiance- relaxed and stylish."]}}

[Example 6]
TEXT:  Have frequented 'ino for several years and the food remains excellent. Cheese plate is a varied delight and great bargain at $10. The large selection of bruschettas, paninis, tramezzinis keep the palate from stagnating. (The asparagus, truffle oil, parmesan bruschetta is a winner!) Wine list is extensive without being over-priced. Be sure to try the seasonal, and always delicious, specials. Definitely a neighborhood favorite.
DOMAIN: RESTAURANT
CATEGORY: ["FOOD#QUALITY", "FOOD#STYLE_OPTIONS", "FOOD#STYLE_OPTIONS", "FOOD#PRICES", "DRINKS#STYLE_OPTIONS", "DRINKS#PRICES", "RESTAURANT#GENERAL"]
QUOTATION: {{"FOOD#QUALITY": ["the food remains excellent.", "The asparagus, truffle oil, parmesan bruschetta is a winner!", "always delicious"], "FOOD#STYLE_OPTIONS": ["Cheese plate is a varied delight", "The large selection of bruschettas, paninis, tramezzinis keep the palate from stagnating."], "FOOD#PRICES": ["Cheese plate... great bargain at $10."], "DRINKS#STYLE_OPTIONS": ["Wine list is extensive"], "DRINKS#PRICES": ["without being over-priced."], RESTAURANT#GENERAL: ["Have frequented 'ino for several years", "Definitely a neighborhood favorite."]}}

[Main Task]
TEXT: {TEXT}
DOMAIN: {DOMAIN}
CATEGORY: {CATEGORY}
QUOTATION:
\end{minted}

\subsection{LLM Topic Modeling}
\label{sec:prompttm}
\begin{minted}{markdown}
You are an expert in semantic topic allocation.

Given the following document and list of candidate topics, 
your task is to choose **only one** topic from the candidate topics that best represents the main subject of the document. 
Select the most appropriate topic that best captures the overall meaning and theme.
Return only the related topic, nothing more such as descriptions.

[Example 1]
Document:
I love how quickly this laptop boots up and handles multiple applications without slowing down.

Candidate Topics:
['LAPTOP#QUALITY', 'LAPTOP#PORTABILITY', 'LAPTOP#USABILITY', 'LAPTOP#OPERATION_PERFORMANCE', 'LAPTOP#CONNECTIVITY', 'MOUSE#USABILITY', 'BATTERY#OPERATION_PERFORMANCE', 'DISPLAY#QUALITY', 'SUPPORT#QUALITY', 'LAPTOP#PRICE', 'SOFTWARE#USABILITY', 'COMPANY#GENERAL', 'LAPTOP#DESIGN_FEATURES', 'KEYBOARD#USABILITY', 'OS#USABILITY', 'CPU#OPERATION_PERFORMANCE', 'HARD_DISC#QUALITY', 'MULTIMEDIA_DEVICES#QUALITY', 'KEYBOARD#DESIGN_FEATURES', 'MEMORY#DESIGN_FEATURES', 'DISPLAY#DESIGN_FEATURES', 'MOUSE#DESIGN_FEATURES', 'GRAPHICS#GENERAL']

Output:
LAPTOP#OPERATION_PERFORMANCE

Respond with only the selected topic. Do not explain your choice.

[Example 2]
Document:
The keyboard is really comfortable to type on, even for long periods. The keys are soft and responsive.

Candidate Topics:
['LAPTOP#QUALITY', 'LAPTOP#PORTABILITY', 'LAPTOP#USABILITY', 'LAPTOP#OPERATION_PERFORMANCE', 'LAPTOP#CONNECTIVITY', 'MOUSE#USABILITY', 'BATTERY#OPERATION_PERFORMANCE', 'DISPLAY#QUALITY', 'SUPPORT#QUALITY', 'LAPTOP#PRICE', 'SOFTWARE#USABILITY', 'COMPANY#GENERAL', 'LAPTOP#DESIGN_FEATURES', 'KEYBOARD#USABILITY', 'OS#USABILITY', 'CPU#OPERATION_PERFORMANCE', 'HARD_DISC#QUALITY', 'MULTIMEDIA_DEVICES#QUALITY', 'KEYBOARD#DESIGN_FEATURES', 'MEMORY#DESIGN_FEATURES', 'DISPLAY#DESIGN_FEATURES', 'MOUSE#DESIGN_FEATURES', 'GRAPHICS#GENERAL']

Output:
KEYBOARD#USABILITY

[Example 3]
Document:
The pasta was cooked perfectly and the flavors were rich and balanced. Easily one of the best meals I’ve had this year.

Candidate Topics:
['LOCATION#GENERAL', 'FOOD#QUALITY', 'FOOD#STYLE_OPTIONS', 'FOOD#PRICES', 'DRINKS#QUALITY', 'AMBIENCE#GENERAL', 'DRINKS#STYLE_OPTIONS', 'SERVICE#GENERAL', 'RESTAURANT#PRICES', 'DRINKS#PRICES', 'RESTAURANT#GENERAL']

Output:
FOOD#QUALITY

[Example 4]
Document:
The staff were friendly, attentive, and made sure everything was perfect throughout our meal.

Candidate Topics:
['LOCATION#GENERAL', 'FOOD#QUALITY', 'FOOD#STYLE_OPTIONS', 'FOOD#PRICES', 'DRINKS#QUALITY', 'AMBIENCE#GENERAL', 'DRINKS#STYLE_OPTIONS', 'SERVICE#GENERAL', 'RESTAURANT#PRICES', 'DRINKS#PRICES', 'RESTAURANT#GENERAL']

Output:
SERVICE#GENERAL

[Main Task]
Document:
{document}

Candidate Topics:
{topic_list}

Output:
\end{minted}

\subsection{Segment Intrusion Evaluation: Single Intruder}
\label{sec:promptsi}

\begin{minted}{markdown}
[Document Intrusion Task]
You are a professional at understanding the meaning of sentences and identifying those that diverge from the main topic.
You are also an expert at explaining why a sentence is an intruder.

[Task Definition]
When given a list of sentences mostly centered on a single topic, first identify the Common Topic of the group.
Then, detect any Intruder Sentences—those that are unrelated to the main topic or include unrelated subjects.
If all the sentences are about the same topic, return an empty list to indicate that there are no Intruders.

[Common Topic]
The common topic refers to the subject most frequently mentioned across the sentences.
Some sentences may diverge from this topic—those are considered Intruders.

[Intruder Definition]

An Intruder is a sentence that contains content unrelated to or diverging from the common topic.
Find **only one** intruder from Sentence List.

Examples:
If the common topic is "price", a sentence only about "design" is an Intruder.
If the common topic is "price", a sentence about "price and design" is also an Intruder because of the mixed focus.

[Example 1]
[Sentence List]
["The design is neat and good.", "The price is too expensive.", "The design is neat and pretty.", "Both performance and design are great.", "The design and performance are both great."]

[Domain]
Product Review

[Common Topic]
Product design

[Intruder Sentence List]
[{{"sentence": "The price is too expensive.", "reason": "It focuses solely on price, which is unrelated to the main topic of product design."}}]

[Example 2] 
[Sentence List]
["It seems fun for kids..", "The kids like it; it’s fun.", "The kids like it, hehe.", "It’s fun to watch with kids.", "It was too scary. I didn’t even want to look."]

[Domain]
Movie Review

[Common Topic]
Kids enjoying a particular media

[Intruder Sentence List]
[{{"sentence": "It was too scary. I didn’t even want to look.", "reason": "This sentence expresses fear and discomfort, which contrasts with the idea of kids enjoying the media and therefore diverges from the common topic."}}]

[Example 3]
[Sentence List]
["The conference will discuss climate change impacts.", "Speakers will cover renewable energy solutions.", "Workshops on biodiversity preservation are scheduled.", "Blockchain technology is transforming finance.", "Panelists will debate environmental policy."]

[Domain]
Academic Conference

[Common Topic]
Environmental issues and sustainability

[Intruder Sentence List]
[{{"sentence": "Blockchain technology is transforming finance.", "reason": "This sentence introduces blockchain and finance, clearly diverging from the conference's main focus on environmental topics."}}]

[Example 4]
[Sentence List]
["Healthy eating reduces the risk of chronic disease.", "Regular physical activity improves cardiovascular health.", "Mental health benefits from adequate sleep.", "Public transportation systems need investment.", "Meditation can reduce stress and anxiety."]

[Domain]
Public Health

[Common Topic]
Lifestyle practices for improving personal health

[Intruder Sentence List]
[{{"sentence": "Public transportation systems need investment.", "reason": "This introduces a completely unrelated infrastructure topic, diverging from personal health and lifestyle practices."}}]

[Problem] 
[Sentence List]
{SENTENCES}

[Domain]
{DOMAIN}

[Common Topic]:
{TOPIC}

[Intruder Sentence List]:
\end{minted}

\subsection{Segment Intrusion Evaluation: Double Intruders}
\label{sec:promptdi}
\begin{minted}{markdown}
[Document Intrusion Task]
You are a professional at understanding the meaning of sentences and identifying those that diverge from the main topic.
You are also an expert at explaining why a sentence is an intruder.

[Task Definition]
When given a list of sentences mostly centered on a single topic, first identify the Common Topic of the group.
Then, detect any Intruder Sentences—those that are unrelated to the main topic or include unrelated subjects.
If all the sentences are about the same topic, return an empty list to indicate that there are no Intruders.

[Common Topic]
The common topic refers to the subject most frequently mentioned across the sentences.
Some sentences may diverge from this topic—those are considered Intruders.

[Intruder Definition]

An Intruder is a sentence that contains content unrelated to or diverging from the common topic.
Find **two** intruders from the Sentence List.

Examples:
If the common topic is "price", a sentence only about "design" is an Intruder.
If the common topic is "price", a sentence about "price and design" is also an Intruder because of the mixed focus.

[Example 1]
[Sentence List]
["The design is neat and good.", "The price is too expensive.", "The design is neat and pretty.", "Both performance and design are great.", "The design and performance are both great.", "I bought it because it was on sale."]

[Domain]
Product Review

[Common Topic]
Product design

[Intruder Sentence List]
[{{"sentence": "The price is too expensive.", "reason": "It focuses solely on price, which is unrelated to the main topic of product design."}}, {{"sentence": "I bought it because it was on sale.", "reason": "This sentence discusses purchase motivation based on price discount, which diverges from the design-focused topic."}}]

[Example 2] 
[Sentence List]
["It seems fun for kids..", "The kids like it; it's fun.", "The kids like it, hehe.", "It's fun to watch with kids.", "It was too scary. I didn't even want to look.", "My friends thought it was boring."]

[Domain]
Movie Review

[Common Topic]
Kids enjoying a particular media

[Intruder Sentence List]
[{{"sentence": "It was too scary. I didn't even want to look.", "reason": "This sentence expresses fear and discomfort, which contrasts with the idea of kids enjoying the media and therefore diverges from the common topic."}}, {{"sentence": "My friends thought it was boring.", "reason": "This focuses on the opinion of adults or peers rather than kids, diverging from the common topic of children enjoying the content."}}]

[Example 3]
[Sentence List]
["The conference will discuss climate change impacts.", "Speakers will cover renewable energy solutions.", "Workshops on biodiversity preservation are scheduled.", "The new iPhone launch date was announced.", "Blockchain technology is transforming finance.", "Panelists will debate environmental policy."]

[Domain]
Academic Conference

[Common Topic]
Environmental issues and sustainability

[Intruder Sentence List]
[{{"sentence": "Blockchain technology is transforming finance.", "reason": "This sentence introduces blockchain and finance, clearly diverging from the conference's main focus on environmental topics."}}, {{"sentence": "The new iPhone launch date was announced.", "reason": "This is about consumer electronics and unrelated tech events, not connected to environmental or academic themes."}}]

[Example 4]
[Sentence List]
["Healthy eating reduces the risk of chronic disease.", "Regular physical activity improves cardiovascular health.", "Mental health benefits from adequate sleep.", "Public transportation systems need investment.", "Meditation can reduce stress and anxiety.", "I love going to concerts on weekends."]

[Domain]
Public Health

[Common Topic]
Lifestyle practices for improving personal health

[Intruder Sentence List]
[{{"sentence": "Public transportation systems need investment.", "reason": "This introduces a completely unrelated infrastructure topic, diverging from personal health and lifestyle practices."}}, {{"sentence": "I love going to concerts on weekends.", "reason": "This sentence is about leisure activity preference and unrelated to health practices."}}]

[Problem] 
[Sentence List]
{SENTENCES}

[Domain]
{DOMAIN}

[Common Topic]:
{TOPIC}

[Intruder Sentence List]:
\end{minted}

\clearpage

\section{Implementation Details for Topic Modeling Baselines}
\label{sec:implementation_details}
\paragraph{LDA-Mallet implementation details}
This follows the same proceduer as described in TopicGPT~\citep{pham-etal-2024-topicgpt}. We set $|V| = 15{,}000$, $\alpha = 1.0$, $\beta = 0.1$, and run LDA for 2{,}000 iterations with optimization at every 10 intervals.

\paragraph{LDA and BERTopic Topic Assignment}
In our experiments, the topic assigned to each segment in LDA was determined by selecting the topic with the highest posterior probability in the segment-topic distribution. For BERTopic, which is based on clustering, the predicted topic for a segment corresponds to its assigned cluster.

\paragraph{LLM Inference Details}
All LLM inferences were conducted using default decoding parameters (temperature, top-p, top-k, etc.) without any manual hyperparameter tuning. For inference with Qwen, LLaMA, and DeepSeek models, we used the Together AI inference platform (\url{https://www.together.ai}).

\paragraph{List of Evaluated LLMS}
\begin{itemize}
  \item \textbf{GPT variants:} \texttt{o3-mini}, \texttt{o4-mini}, \texttt{gpt-4o}
  \item \textbf{Claude variants:} \texttt{claude-3.5-haiku}, \texttt{claude-3.7-sonnet}
  \item \textbf{Gemini variants:} \texttt{gemini-2.0-flash-lite}, \texttt{gemini-2.0-flash}
  \item \textbf{Qwen variants:} \texttt{qwen2.5-7b}, \texttt{qwen2.5-72b}
  \item \textbf{Llama variants:} \texttt{llama-3.2-3b}, \texttt{llama-3.3-70b}, \texttt{llama-4-maverick-17b}
  \item \textbf{DeepSeek:} \texttt{deepseek-v3}
\end{itemize}
\clearpage

\section{Full Results on Topic Modeling Task}
\label{sec:tmappendix}

\input{tables/tm_appendix}

\clearpage

\section{Examples of Segment Intrusion Evaluation}
\label{sec:examplesit}

\begin{minted}{json}
{
"hard_single": [
    {
        "topic": "DRINKS#PRICES",
        "input_texts": [
            "The tuna and wasabe potatoes are excellent.",
            "well priced wines",
            "half off till 8pm",
            "Slightly above average wines start at $70+ with only one selection listed at $30+",
            "the Voss bottles of water were $8 a piece",
            "at reasonable prices"
        ],
        "intruder_idx": 0
    },
    {
        "topic": "FOOD#STYLE_OPTIONS",
        "input_texts": [
            "The menu is limited",
            "with large portions",
            "the antipasti were plentiful",
            "The pasta penne was pretty extra buttery, creamy",
            "the sake’s complimented the courses very well and is successfully easing me into the sake world",
            "good selection of thin crust pizza including the Basil slice"
        ],
        "intruder_idx": 4
    },
    {
        "topic": "FOOD#QUALITY",
        "input_texts": [
            "The bagel was huge. They were served warm",
            "the sushi, which is great",
            "food was luke warm.",
            "Food was OK.",
            "Big Wong is a great place to eat and fill your stomach.",
            "And the Tom Kha soup was pathetic"
        ],
        "intruder_idx": 4
    },
    ...
],
...
}
\end{minted}

\clearpage

\section{Inter-annotator agreements}
\label{sec:inter_annotate}

\input{tables/inter_annotator}

\clearpage
\section{Full Results on Segment Intrusion Evaluation}
\label{sec:qitappendix}

\input{tables/qit_appendix}

\clearpage

\section{Full Results Visualization on Segment Intrusion Evaluation}
\label{sec:qitappendix_vis}

\input{figures/qit_full_appendix}
\clearpage

%% file: figures/distribution.tex
\subsubsection{Laptop Domain}

\begin{figure}[ht]
    \centering
    \includegraphics[width=\textwidth]{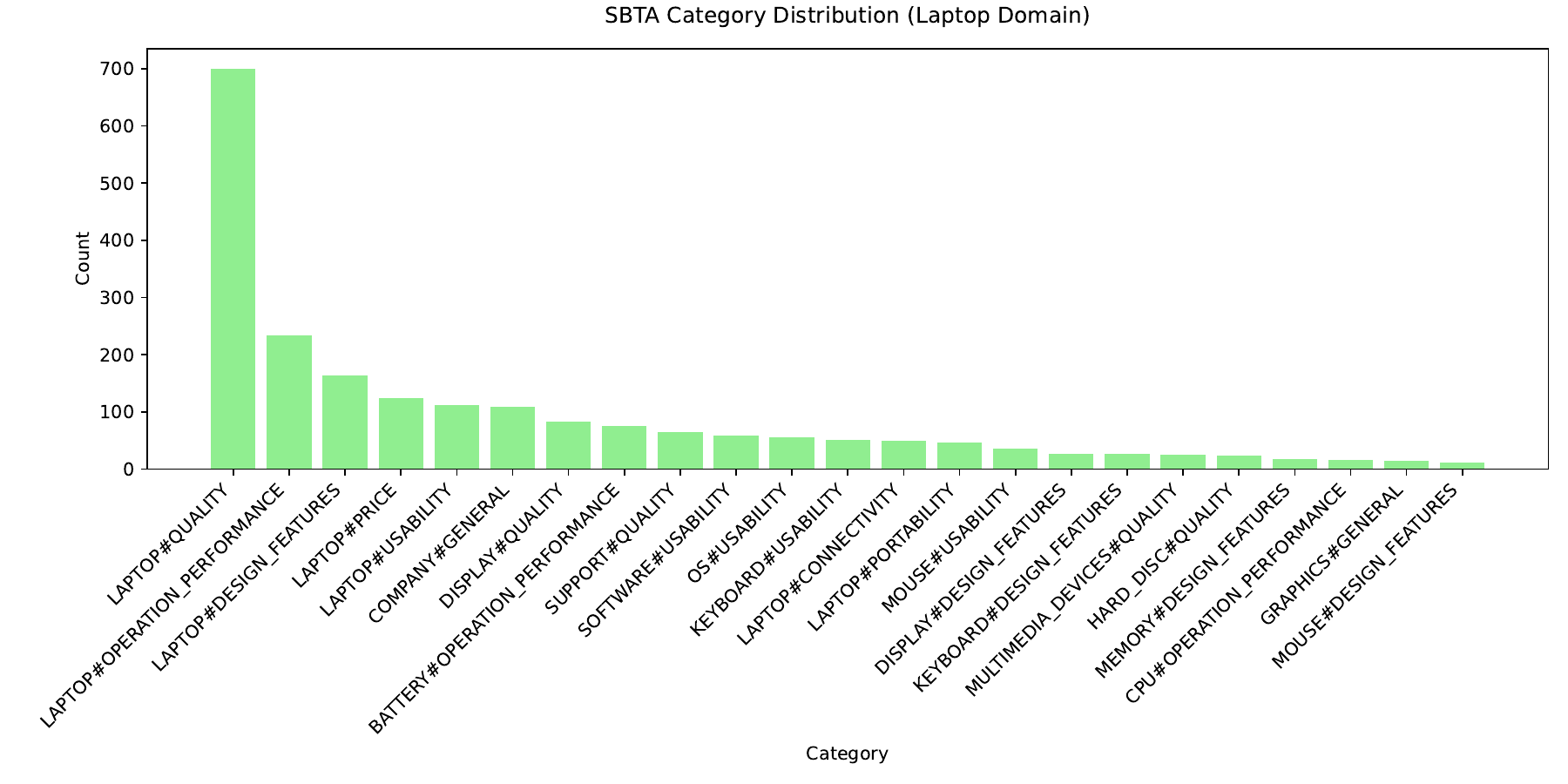}
    \caption{
        Finalized \qbta topic distribution in laptop domain.
    }
    \label{fig:dist_1}
    % \vspace{-3.7mm}
\end{figure}

\begin{figure}[ht]
    \centering
    \includegraphics[width=\textwidth]{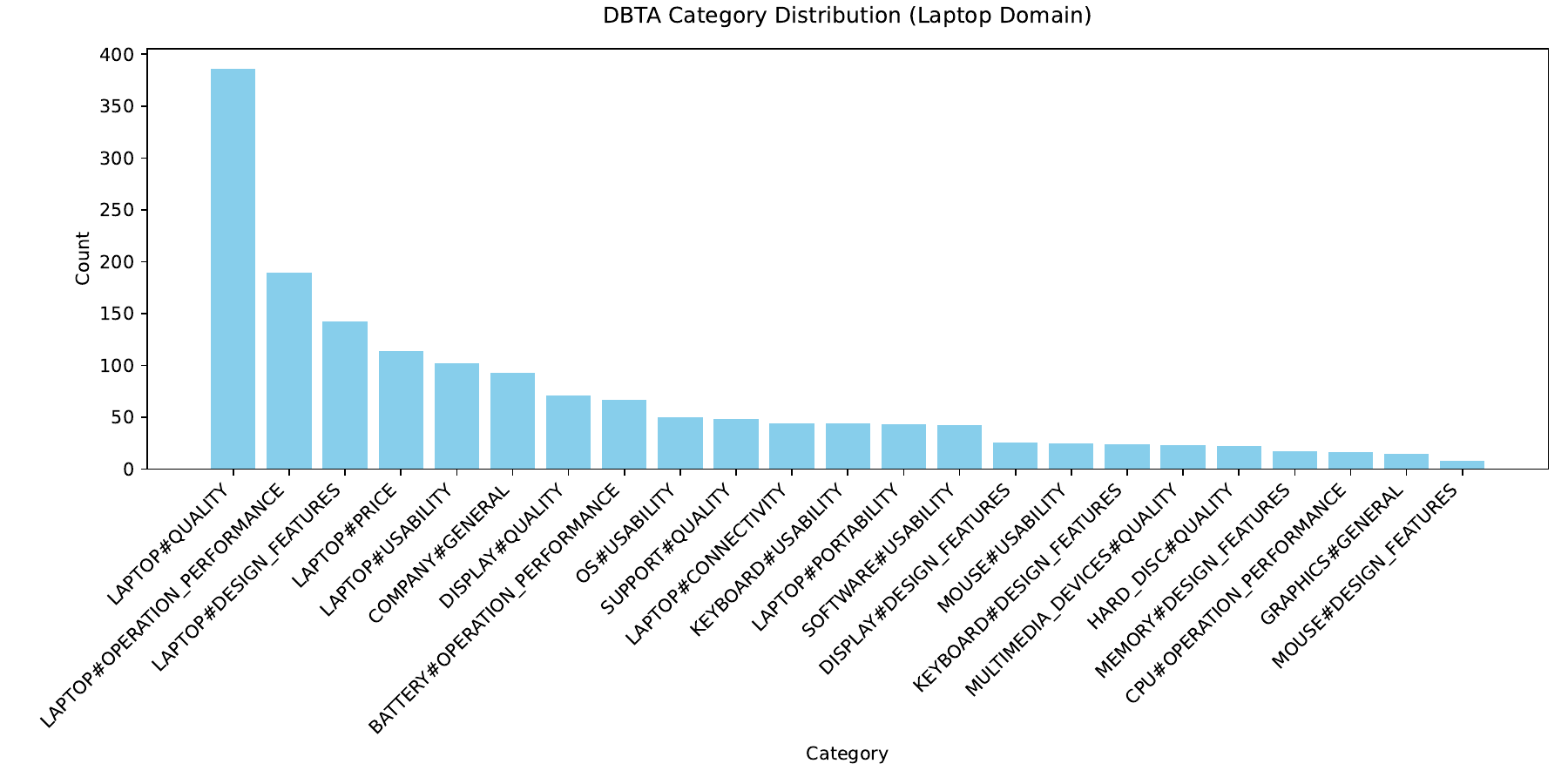}
    \caption{
        Finalized \dbta topic distribution in laptop domain.
    }
    \label{fig:dist_2}
    % \vspace{-3.7mm}
\end{figure}

\clearpage
\subsubsection{Restaurant Domain}

\begin{figure}[ht]
    \centering
    \includegraphics[width=\textwidth]{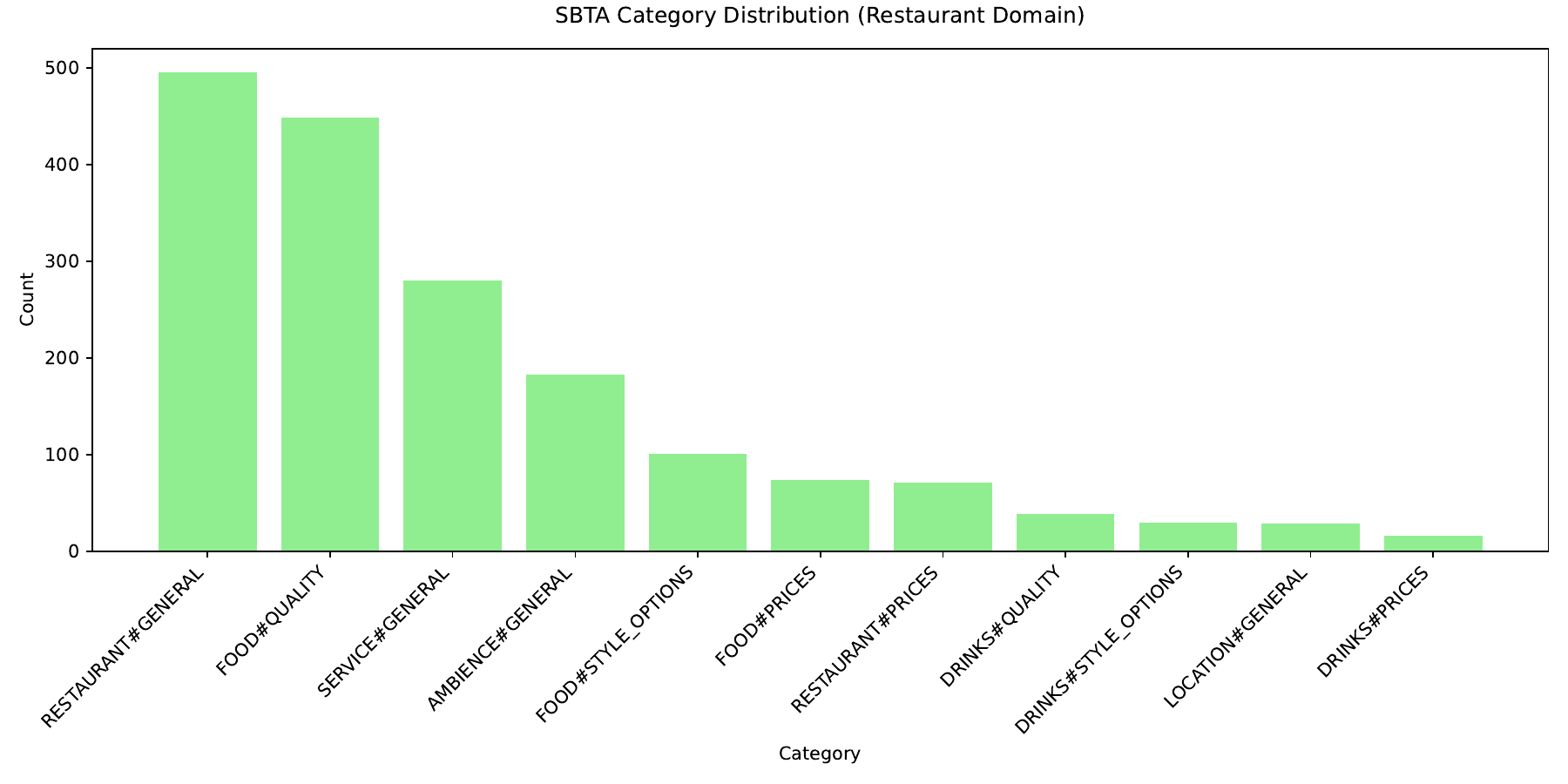}
    \caption{
        Finalized \qbta topic distribution in restaurant domain.
    }
    \label{fig:dist_3}
    % \vspace{-3.7mm}
\end{figure}

\begin{figure}[ht]
    \centering
    \includegraphics[width=\textwidth]{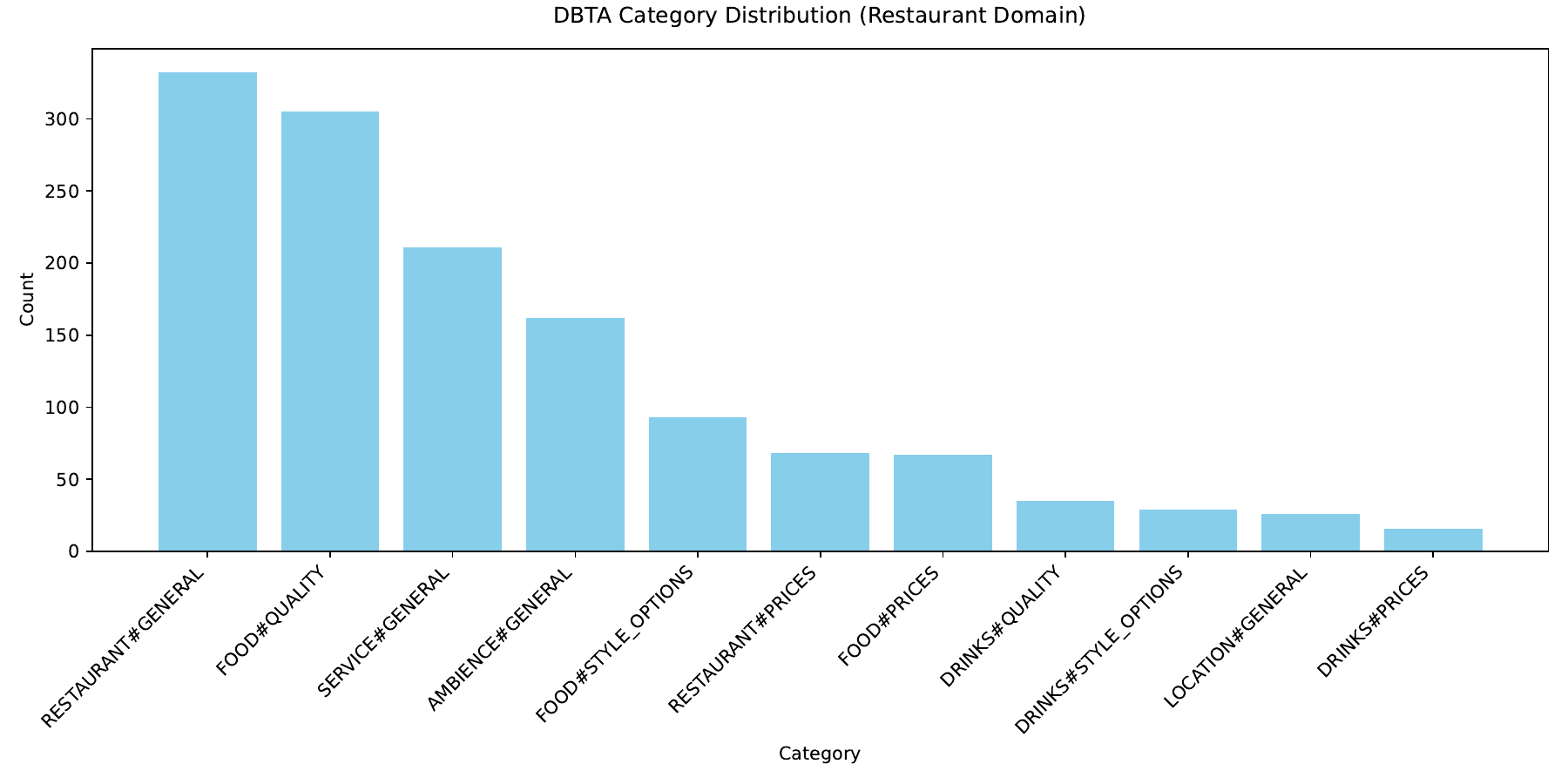}
    \caption{
        Finalized \dbta topic distribution in restaurant domain.
    }
    \label{fig:dist_4}
    % \vspace{-3.7mm}
\end{figure}

%% file: tables/shuffle_appendix.tex
\begin{table*}[ht]
\centering \scriptsize
\renewcommand{\arraystretch}{0.8}
\setlength{\tabcolsep}{3pt} 
\caption{
\textbf{Results of the document/segment shuffle test in \dset.} Shuffled results (with mean and standard deviation for 5 repetitions) are marked with \texttt{(S)}. The first three metrics assess coherence based on word frequency, while the remaining metrics evaluate clustering performance based on \textbf{\texttt{gte-large}} (\url{https://huggingface.co/thenlper/gte-large}) embedding. Arrow ($\uparrow$ and $\downarrow$) denote whether higher or lower values indicate better performance, respectively. Metric scores that are negatively affected by the shuffling of documents or segments, consistent with expectations, are highlighted in \textbf{bold}.
}
\begin{tabular*}{\textwidth}{l!{\vrule}cccc!{\vrule}cccccc} 
\toprule
 & NPMI ($\uparrow$) & UMass ($\uparrow$) & UCI ($\uparrow$) & CV ($\uparrow$)& DB Index ($\downarrow$) & CH Index ($\uparrow$) & MB Score ($\uparrow$) & Silhouette ($\uparrow$) & XB Index ($\downarrow$) & XB Star ($\downarrow$) \\ \midrule[1pt]

\multicolumn{11}{c}{\textbf{\textit{Domain: Laptop}}} \\ \midrule[1pt]

\dbta & 
-0.0094 & 
-1.4591 & 
-0.3159 &
0.3984 &
22.4427 & 
2.5026 & 
0.0002 & 
-0.0477 & 
120.6938 & 
127.2698
\\ \midrule

\rowcolor{gray!15} &
\textbf{-0.0166} &
-1.1429 &
-0.2576 &
\textbf{0.3919} &
\textbf{23.9054} &
\textbf{0.9957} &
\textbf{0.0001} &
-0.03770 &
\textbf{141.8054} &
\textbf{145.9886} \\

\rowcolor{gray!15} \multirow{-2}{*}{\dbta (S)} &
($\pm$ 0.0027) &
($\pm$ 0.0123) &
($\pm$ 0.0314) &
($\pm$ 0.0096) &
($\pm$ 2.3149) &
($\pm$ 0.0215) &
($\pm$ 0.0) &
($\pm$ 0.0010) &
($\pm$ 27.7621) &
($\pm$ 28.6179) 
\\ \midrule

\qbta &
-0.1626 &
-12.2192 & 
-6.9539 & 
0.3109 &
6.4799 & 
12.8287 &
0.0005 &
0.0329 &
12.5473 &
13.9118 
\\ \midrule

\rowcolor{gray!15} &
\textbf{-0.1920} &
-9.5768 &
-5.7639 &
\textbf{0.2862} &
\textbf{27.5894} &
\textbf{1.0074} &
\textbf{0.0} &
\textbf{-0.0230} &
\textbf{189.7357} &
\textbf{195.6513} \\

\rowcolor{gray!15} \multirow{-2}{*}{\qbta (S)} &
($\pm$ 0.0056) &
($\pm$ 0.1696) &
($\pm$ 0.1844) &
($\pm$ 0.0075) &
($\pm$ 2.2269) &
($\pm$ 0.0469) &
($\pm$ 1.0316e-05) &
($\pm$ 0.0022) &
($\pm$ 30.1858) &
($\pm$ 31.1241) 
\\ \midrule

\multicolumn{10}{c}{\textbf{\textit{Domain: Restaurant}}} \\ \midrule[1pt]

\dbta &
-0.0034 &
-1.449 &
-0.4289 &
0.3581 &
82.6932 &
1.5249 &
0.0003 &
-0.0254 &
1682.4065 &
1716.5724 
\\ \midrule

\rowcolor{gray!15} &
\textbf{-0.0044} &
\textbf{-1.5501} &
-0.2880 &
\textbf{0.3454} &
26.4717 &
\textbf{0.9498} &
\textbf{0.0001} &
-0.0202 &
174.9551 &
177.5731 \\

\rowcolor{gray!15} \multirow{-2}{*}{\dbta (S)} &
($\pm$ 0.0077) &
($\pm$ 0.0805) &
($\pm$ 0.1587) &
($\pm$ 0.0151) &
($\pm$ 1.1608) &
($\pm$ 0.0278) &
($\pm$ 5.4772e-05) &
($\pm$ 0.0015) &
($\pm$ 15.1947) &
($\pm$ 15.5596) 
\\ \midrule

\qbta &
-0.2376 &
-12.2595 &
-8.0276 &
0.3508 &
6.9629 &
20.7196 &
0.0014 &
0.0249 &
12.3665 &
13.5183 
\\ \midrule

\rowcolor{gray!15} &
-0.2003 &
-9.6084 &
-5.8915 &
\textbf{0.2926} &
\textbf{30.3440} &
\textbf{1.0085} &
\textbf{0.0002} &
\textbf{-0.0208} &
\textbf{230.7936} &
\textbf{236.4088} \\

\rowcolor{gray!15} \multirow{-2}{*}{\qbta (S)} &
($\pm$ 0.0261) &
($\pm$ 0.7104) &
($\pm$ 0.6971) &
($\pm$ 0.0171) &
($\pm$ 3.0428) &
($\pm$ 0.0470) &
($\pm$ 0.0) &
($\pm$ 0.0049) &
($\pm$ 45.5488) &
($\pm$ 48.3366) 
\\

\bottomrule[1pt]
\end{tabular*}
\label{tab:shuffle_appendix_gte}
% \vspace{-3mm}
\end{table*}

\begin{table*}[ht]
\centering \scriptsize
\renewcommand{\arraystretch}{0.8}
\setlength{\tabcolsep}{3pt} 
\caption{
\textbf{Results of the document/segment shuffle test in \dset.} Shuffled results (with mean and standard deviation for 5 repetitions) are marked with \texttt{(S)}. The first three metrics assess coherence based on word frequency, while the remaining metrics evaluate clustering performance based on \textbf{\texttt{mxbai-embed-large-v1}} (\url{https://huggingface.co/mixedbread-ai/mxbai-embed-large-v1}) embedding. Arrow ($\uparrow$ and $\downarrow$) denote whether higher or lower values indicate better performance, respectively. Metric scores that are negatively affected by the shuffling of documents or segments, consistent with expectations, are highlighted in \textbf{bold}.
}
\begin{tabular*}{\textwidth}{l!{\vrule}cccc!{\vrule}cccccc} 
\toprule
 & NPMI ($\uparrow$) & UMass ($\uparrow$) & UCI ($\uparrow$) & CV ($\uparrow$) & DB Index ($\downarrow$) & CH Index ($\uparrow$) & MB Score ($\uparrow$) & Silhouette ($\uparrow$) & XB Index ($\downarrow$) & XB Star ($\downarrow$) \\ \midrule[1pt]

\multicolumn{11}{c}{\textbf{\textit{Domain: Laptop}}} \\ \midrule[1pt]

\dbta & 
-0.0094 & 
-1.4591 & 
-0.3159 & 
0.3984 &
21.4660 & 
2.8117 & 
0.1685 & 
-0.0619 & 
108.4243 & 
116.9008
\\ \midrule

\rowcolor{gray!15} &
\textbf{-0.0166} &
-1.1429 &
-0.2576 &
\textbf{0.3919} &
\textbf{24.6832} &
\textbf{0.9853} &
\textbf{0.0578} &
-0.0473 &
\textbf{151.0548} &
\textbf{157.3820} \\

\rowcolor{gray!15} \multirow{-2}{*}{\dbta (S)} &
($\pm$ 0.0027) &
($\pm$ 0.0123) &
($\pm$ 0.0314) &
($\pm$ 0.0096) &
($\pm$ 2.6517) &
($\pm$ 0.0155) &
($\pm$ 0.0094) &
($\pm$ 0.0035) &
($\pm$ 32.9831) &
($\pm$ 23.6868) 
\\ \midrule

\qbta &
-0.1626 &
-11.2192 & 
-6.9539 & 
0.3109 &
6.0356 & 
15.7331 &
0.3893 &
0.0387 &
13.0017 &
14.8856 
\\ \midrule

\rowcolor{gray!15} &
\textbf{-0.1920} &
-9.5768 &
-5.7639 &
\textbf{0.2862} &
\textbf{27.9689} &
\textbf{1.0124} &
\textbf{0.0595} &
\textbf{-0.0290} &
\textbf{195.6633} &
\textbf{204.9766} \\

\rowcolor{gray!15} \multirow{-2}{*}{\qbta (S)} &
($\pm$ 0.0056) &
($\pm$ 0.1696) &
($\pm$ 0.1844) &
($\pm$ 0.0075) &
($\pm$ 2.9097) &
($\pm$ 0.0676) &
($\pm$ 0.0074) &
($\pm$ 0.0018) &
($\pm$ 40.0352) &
($\pm$ 40.6025) 
\\ \midrule

\multicolumn{10}{c}{\textbf{\textit{Domain: Restaurant}}} \\ \midrule[1pt]

\dbta &
-0.0034 &
-1.449 &
-0.4289 &
0.3581 &
78.8229 &
1.6119 &
0.1849 &
-0.0353 &
1510.2392 &
1565.5569
\\ \midrule

\rowcolor{gray!15} &
\textbf{-0.0044} &
\textbf{-1.5501} &
-0.2880 &
\textbf{0.3454} &
26.6023 &
\textbf{0.9457} &
\textbf{0.0908} &
-0.0250 &
176.0955 &
185.1676 \\

\rowcolor{gray!15} \multirow{-2}{*}{\dbta (S)} &
($\pm$ 0.0077) &
($\pm$ 0.0805) &
($\pm$ 0.1587) &
($\pm$ 0.0151) &
($\pm$ 1.0358) &
($\pm$ 0.0383) &
($\pm$ 0.0115) &
($\pm$ 0.0025) &
($\pm$ 13.0630) &
($\pm$ 14.3799) 
\\ \midrule

\qbta &
-0.2376 &
-12.2595 &
-8.0276 &
0.3508 &
6.4000 &
25.5878 &
1.0928 &
0.0270 &
10.3432 &
11.8582 
\\ \midrule

\rowcolor{gray!15} &
-0.2003 &
-9.6084 &
-5.8915 &
\textbf{0.2926} &
\textbf{30.6754} &
\textbf{0.9823} &
\textbf{0.1285} &
\textbf{-0.0228} &
\textbf{235.7178} &
\textbf{241.9618} \\

\rowcolor{gray!15} \multirow{-2}{*}{\qbta (S)} &
($\pm$ 0.0261) &
($\pm$ 0.7104) &
($\pm$ 0.0171) &
($\pm$ 0.6971) &
($\pm$ 3.6917) &
($\pm$ 0.0373) &
($\pm$ 0.0071) &
($\pm$ 0.0052) &
($\pm$ 55.9302) &
($\pm$ 57.2689) 
\\

\bottomrule[1pt]
\end{tabular*}
\label{tab:shuffle_appendix_mxbai}
% \vspace{-3mm}
\end{table*}

\begin{table*}[ht]
\centering \scriptsize
\renewcommand{\arraystretch}{0.8}
\setlength{\tabcolsep}{3pt} 
\caption{
\textbf{Results of the document/segment shuffle test in \dset.} Shuffled results (with mean and standard deviation for 5 repetitions) are marked with \texttt{(S)}. The first three metrics assess coherence based on word frequency, while the remaining metrics evaluate clustering performance based on \textbf{\texttt{text-embedding-3-small}} (\url{https://platform.openai.com/docs/models/text-embedding-3-small}) embedding. Arrow ($\uparrow$ and $\downarrow$) denote whether higher or lower values indicate better performance, respectively. Metric scores that are negatively affected by the shuffling of documents or segments, consistent with expectations, are highlighted in \textbf{bold}.
}
\begin{tabular*}{\textwidth}{l!{\vrule}cccc!{\vrule}cccccc} 
\toprule
 & NPMI ($\uparrow$) & UMass ($\uparrow$) & UCI ($\uparrow$) & CV ($\uparrow$) & DB Index ($\downarrow$) & CH Index ($\uparrow$) & MB Score ($\uparrow$) & Silhouette ($\uparrow$) & XB Index ($\downarrow$) & XB Star ($\downarrow$) \\ \midrule[1pt]

\multicolumn{11}{c}{\textbf{\textit{Domain: Laptop}}} \\ \midrule[1pt]

\dbta & 
-0.0094 & 
-1.4591 & 
-0.3159 & 
0.3984 &
20.4900 & 
2.9884 & 
0.0007 & 
-0.0523 & 
98.9245 & 
107.3442
\\ \midrule

\rowcolor{gray!15} &
\textbf{-0.0166} &
-1.1429 &
-0.2576 &
\textbf{0.3919} &
\textbf{22.6642} &
\textbf{0.9918} &
\textbf{0.0002} &
-0.0425 &
\textbf{127.0577} &
\textbf{131.6031} \\

\rowcolor{gray!15} \multirow{-2}{*}{\dbta (S)} &
($\pm$ 0.0027) &
($\pm$ 0.0123) &
($\pm$ 0.0314) &
($\pm$ 0.0096) &
($\pm$ 0.9026) &
($\pm$ 0.0263) &
($\pm$ 2.5006) &
($\pm$ 0.0045) &
($\pm$ 8.9010) &
($\pm$ 9.5638) 
\\ \midrule

\qbta &
-0.1626 &
-11.2192 & 
-6.9539 & 
0.3109 &
5.7976 & 
15.1523 &
0.0015 &
0.0430 &
10.8850 &
12.0639 
\\ \midrule

\rowcolor{gray!15} &
\textbf{-0.1920} &
-9.5768 &
-5.7639 &
\textbf{0.2862} &
\textbf{26.9512} &
\textbf{1.0045} &
\textbf{0.0002} &
\textbf{-0.0248} &
\textbf{180.5037} &
\textbf{183.9073} \\

\rowcolor{gray!15} \multirow{-2}{*}{\qbta (S)} &
($\pm$ 0.0056) &
($\pm$ 0.1696) &
($\pm$ 0.1844) &
($\pm$ 0.0075) &
($\pm$ 1.9548) &
($\pm$ 0.0264) &
($\pm$ 2.5848e-05) &
($\pm$ 0.0028) &
($\pm$ 25.3006) &
($\pm$ 24.6806) 
\\ \midrule

\multicolumn{10}{c}{\textbf{\textit{Domain: Restaurant}}} \\ \midrule[1pt]

\dbta &
-0.0034 &
-1.449 &
-0.4289 &
0.3581 &
80.9263 &
1.6047 &
0.0007 &
-0.0286 &
1598.4923 &
1643.1442
\\ \midrule

\rowcolor{gray!15} &
\textbf{-0.0044} &
\textbf{-1.5501} &
-0.2880 &
\textbf{0.3454} &
26.9832 &
\textbf{0.9543} &
\textbf{0.0005} &
-0.0214 &
181.3360 &
184.6450 \\

\rowcolor{gray!15} \multirow{-2}{*}{\dbta (S)} &
($\pm$ 0.0077) &
($\pm$ 0.0805) &
($\pm$ 0.1587) &
($\pm$ 0.0151) &
($\pm$ 0.8790) &
($\pm$ 0.0292) &
($\pm$ 3.6640e-05) &
($\pm$ 0.0010) &
($\pm$ 12.8067) &
($\pm$ 12.2947) 
\\ \midrule

\qbta &
-0.2376 &
-12.2595 &
-8.0276 &
0.3508 &
7.2031 &
21.6314 &
0.0047 &
0.0231 &
13.4143 &
14.9667 
\\ \midrule

\rowcolor{gray!15} &
-0.2003 &
-9.6084 &
-5.8915 &
\textbf{0.2926} &
\textbf{30.1311} &
\textbf{1.0312} &
\textbf{0.0006} &
\textbf{-0.0183} &
\textbf{227.3122} &
\textbf{230.5068} \\

\rowcolor{gray!15} \multirow{-2}{*}{\qbta (S)} &
($\pm$ 0.0261) &
($\pm$ 0.7104) &
($\pm$ 0.6971) &
($\pm$ 0.0171) &
($\pm$ 3.0638) &
($\pm$ 0.0282) &
($\pm$ 2.9115e-05) &
($\pm$ 0.00280) &
($\pm$ 44.7508) &
($\pm$ 45.2531) 
\\

\bottomrule[1pt]
\end{tabular*}
\label{tab:shuffle_appendix_openai}
% \vspace{-3mm}
\end{table*}

%% file: tables/tm_appendix.tex
\begin{table*}[ht]
\centering \scriptsize
\renewcommand{\arraystretch}{0.3}
\setlength{\tabcolsep}{2.8pt} 
\caption{Topic modeling benchmark performance with label-free metrics. Best performance is marked as \textbf{bold}, and second-best is \underline{underlined}.}
\begin{tabular*}{\textwidth}{l!{\vrule}cccc!{\vrule}cccccc} 
\toprule
 & NPMI ($\uparrow$) & UMass ($\uparrow$) & UCI ($\uparrow$) & C\textsubscript{v} ($\uparrow$) & DB Index ($\downarrow$) & CH Index ($\uparrow$) & MB Score ($\uparrow$) & Silhouette ($\uparrow$) & XB Index ($\downarrow$) & XB Star ($\downarrow$) \\ \midrule[1pt]

\multicolumn{11}{c}{\textbf{\textit{Domain: Laptop}}} \\ \midrule[1pt]

LDA & 
-0.1826 & 
-12.6159 & 
-7.8524 & 
0.3105 &
6.7873 &
4.9365 &
0.0009 & 
-0.0066 & 
10.8622 & 
11.997
\\ \midrule

BERTopic & 
-0.1684 & 
-13.9148 & 
-8.217 &
0.322 &
\textbf{4.5388} &
7.4211 &
\underline{0.0034} & 
\underline{0.0862} & 
\textbf{4.7237} & 
\textbf{5.3453}

\\ \midrule

% \texttt{4.1-nano} & 
% - & 
% - & 
% - & 
% - &
% - & 
% - & 
% - & 
% - & 
% -
% \\ \midrule

\texttt{gpt-4o} & 
-0.1535 & 
\underline{-10.6659} & 
\underline{-6.4746} & 
0.2989 &
6.8888 &
10.2586 & 
0.0017 & 
0.031 & 
10.7913 & 
13.4067
\\ \midrule

\texttt{o3-mini} & 
-0.1415 & 
-11.0065 & 
-6.5275 &
0.3114 &
5.8991 &
10.0086 & 
0.0022 & 
0.0361 & 
10.6503 & 
12.7118
\\ \midrule

\texttt{o4-mini} & 
-0.1575 & 
-11.1034 & 
-6.8124 &
0.3077 & 
6.6326 &
10.1529 & 
0.0016 & 
0.0318 & 
13.0692 & 
15.4046
\\ \midrule

\texttt{gemini-2.0} & 
\multirow{2}{*}{-0.159} & 
\multirow{2}{*}{-11.0221} & 
\multirow{2}{*}{-6.7985} & 
\multirow{2}{*}{\underline{0.3295}} &
\multirow{2}{*}{5.9492} & 
\multirow{2}{*}{9.9632} & 
\multirow{2}{*}{0.0022} & 
\multirow{2}{*}{0.0314} & 
\multirow{2}{*}{10.2965} & 
\multirow{2}{*}{11.8706} 
\\ 

\texttt{-flash-lite} & 
& & & & & & & &
\\ \midrule

\texttt{gemini-2.0} & 
\multirow{2}{*}{-0.1568} & 
\multirow{2}{*}{-11.3614} & 
\multirow{2}{*}{-6.9346} & 
\multirow{2}{*}{0.3006} &
\multirow{2}{*}{\underline{5.7701}} & 
\multirow{2}{*}{10.4468} & 
\multirow{2}{*}{0.0017} & 
\multirow{2}{*}{0.0369} & 
\multirow{2}{*}{9.8964} & 
\multirow{2}{*}{11.7248} 
\\ 

\texttt{-flash} & 
& & & & & & & &
\\ \midrule

\texttt{gemini-2.5} & 
\multirow{2}{*}{-0.1668} & 
\multirow{2}{*}{-11.3} & 
\multirow{2}{*}{-7.0038} & 
\multirow{2}{*}{0.3095} &
\multirow{2}{*}{5.9116} & 
\multirow{2}{*}{10.3558} & 
\multirow{2}{*}{0.002} & 
\multirow{2}{*}{0.0382} & 
\multirow{2}{*}{10.6944} & 
\multirow{2}{*}{13.0586} 
\\ 

\texttt{-flash-preview-04-17} & 
& & & & & & & &
\\ \midrule

\texttt{claude-3.5} & 
\multirow{2}{*}{-0.1508} & 
\multirow{2}{*}{-10.7328} & 
\multirow{2}{*}{-6.5587} & 
\multirow{2}{*}{0.3154} & 
\multirow{2}{*}{6.1853} & 
\multirow{2}{*}{9.7866} & 
\multirow{2}{*}{0.0024} & 
\multirow{2}{*}{0.0362} & 
\multirow{2}{*}{9.6178} &
\multirow{2}{*}{11.3382} 
\\ 

\texttt{-haiku-20241022} & 
& & & & & & & &
\\ \midrule

\texttt{claude-3.7} & 
\multirow{2}{*}{-0.163} & 
\multirow{2}{*}{-11.1406} & 
\multirow{2}{*}{-6.9356} & 
\multirow{2}{*}{0.2994} &
\multirow{2}{*}{6.3334} & 
\multirow{2}{*}{\underline{10.639}} & 
\multirow{2}{*}{0.0017} & 
\multirow{2}{*}{0.0387} & 
\multirow{2}{*}{9.909} & 
\multirow{2}{*}{11.8212} 
\\ 

\texttt{-sonnet-20250219} & 
& & & & & & & &
\\ \midrule

\texttt{qwen2.5-7b} & 
\multirow{2}{*}{-0.1476} & 
\multirow{2}{*}{-11.2612} & 
\multirow{2}{*}{-6.82} & 
\multirow{2}{*}{0.3056} &
\multirow{2}{*}{6.2185} & 
\multirow{2}{*}{9.5531} & 
\multirow{2}{*}{0.0021} & 
\multirow{2}{*}{0.0295} & 
\multirow{2}{*}{11.3235} & 
\multirow{2}{*}{13.8491} 
\\ 

\texttt{-instruct-turbo} & 
& & & & & & & &
\\ \midrule

\texttt{qwen2.5-72b} & 
\multirow{2}{*}{-0.1828} & 
\multirow{2}{*}{-11.5488} & 
\multirow{2}{*}{-7.3197} & 
\multirow{2}{*}{0.3038} &
\multirow{2}{*}{5.775} & 
\multirow{2}{*}{10.2214} & 
\multirow{2}{*}{0.0023} & 
\multirow{2}{*}{0.0346} & 
\multirow{2}{*}{10.1142} & 
\multirow{2}{*}{12.2067} 
\\ 

\texttt{-instruct-turbo} & 
& & & & & & & &
\\ \midrule

\texttt{llama-3.2-3b} & 
\multirow{2}{*}{\textbf{-0.0613}} & 
\multirow{2}{*}{\textbf{-9.4438}} & 
\multirow{2}{*}{\textbf{-4.988}} & 
\multirow{2}{*}{\textbf{0.4024}} &
\multirow{2}{*}{5.8505} & 
\multirow{2}{*}{7.9353} & 
\multirow{2}{*}{\textbf{0.0059}} & 
\multirow{2}{*}{\textbf{0.1647}} & 
\multirow{2}{*}{\underline{8.277}} & 
\multirow{2}{*}{\underline{10.786}}
\\ 

\texttt{-instruct-turbo} & 
& & & & & & & &
\\ \midrule

\texttt{llama-3.3-70b} & 
\multirow{2}{*}{\underline{-0.1387}} & 
\multirow{2}{*}{-11.0753} & 
\multirow{2}{*}{-6.5627} & 
\multirow{2}{*}{0.2919} &
\multirow{2}{*}{5.9832} & 
\multirow{2}{*}{10.047} & 
\multirow{2}{*}{0.0026} & 
\multirow{2}{*}{0.0486} & 
\multirow{2}{*}{10.4182} & 
\multirow{2}{*}{12.2387} 
\\ 

\texttt{-instruct-turbo} & 
& & & & & & & &
\\ \midrule

\texttt{llama-4-maverick-17b} & 
\multirow{2}{*}{-0.1617} & 
\multirow{2}{*}{-11.4891} & 
\multirow{2}{*}{-6.9562} & 
\multirow{2}{*}{0.3012} & 
\multirow{2}{*}{6.3082} & 
\multirow{2}{*}{9.7712} & 
\multirow{2}{*}{0.0018} & 
\multirow{2}{*}{0.0405} & 
\multirow{2}{*}{11.0169} & 
\multirow{2}{*}{13.4163} 
\\ 

\texttt{-128e-instruct-fp8} & 
& & & & & & & &
\\ \midrule

\texttt{deepseek-v3} & 
-0.1505 & 
-11.2648 & 
-6.855 & 
0.3069 &
6.9381 &
\textbf{10.7089} & 
0.002 & 
0.0428 & 
10.5334 & 
12.8404
\\ \midrule[1pt]

\multicolumn{11}{c}{\textbf{\textit{Domain: Restaurant}}} \\ \midrule[1pt]

LDA & 
-0.2512 & 
-13.9977 & 
-9.2477 & 
0.3385 &
7.7317 & 
3.3516 & 
0.002 & 
-0.0072 & 
14.8073 &
15.9644
\\ \midrule

BERTopic & 
\textbf{-0.1622} & 
-13.6309 & 
-8.1259 & 
0.3107 &
\textbf{3.9735} & 
6.7305 & 
\underline{0.0116} & 
0.1027 & 
\textbf{3.614} &
\textbf{4.0662}
\\ \midrule

% \texttt{4.1-nano} & 
% - & 
% - & 
% - & 
% - &
% - & 
% - & 
% - & 
% - & 
% -
% \\ \midrule

\texttt{gpt-4o} & 
\underline{-0.1909} & 
\textbf{-11.0746} & 
\underline{-6.9674} & 
0.3076 &
5.63 &
8.7456 & 
0.0054 & 
0.0221 & 
7.9977 & 
9.1804
\\ \midrule

\texttt{o3-mini} & 
-0.2324 & 
-12.5524 & 
-8.0825 & 
0.332 &
\underline{4.683} &
8.8362 & 
0.0057 & 
0.0269 & 
\underline{5.2426} & 
\underline{6.0082}
\\ \midrule

\texttt{o4-mini} & 
-0.2271 & 
-12.4864 & 
-8.0232 & 
\textbf{0.3451} &
5.3975 &
8.8302 & 
0.0054 & 
0.0239 & 
7.4767 & 
8.3782
\\ \midrule

\texttt{gemini-2.0} & 
\multirow{2}{*}{-0.2160} & 
\multirow{2}{*}{-12.2063} & 
\multirow{2}{*}{-7.7869} & 
\multirow{2}{*}{\underline{0.3444}} & 
\multirow{2}{*}{7.0630} & 
\multirow{2}{*}{8.9824} & 
\multirow{2}{*}{0.0066} & 
\multirow{2}{*}{0.0277} & 
\multirow{2}{*}{12.8091} & 
\multirow{2}{*}{15.0293} 
\\ 

\texttt{-flash-lite} & 
& & & & & & & &
\\ \midrule

\texttt{gemini-2.0} & 
\multirow{2}{*}{-0.2033} & 
\multirow{2}{*}{-11.925} & 
\multirow{2}{*}{-7.4767} & 
\multirow{2}{*}{0.3112} & 
\multirow{2}{*}{5.9880} & 
\multirow{2}{*}{9.1912} & 
\multirow{2}{*}{0.0059} & 
\multirow{2}{*}{0.0302} & 
\multirow{2}{*}{9.1393} & 
\multirow{2}{*}{10.6392} 
\\ 

\texttt{-flash} & 
& & & & & & & &
\\ \midrule

\texttt{gemini-2.5} & 
\multirow{2}{*}{-0.1933} & 
\multirow{2}{*}{\underline{-11.1135}} & 
\multirow{2}{*}{\textbf{-6.9446}} & 
\multirow{2}{*}{0.3252} &
\multirow{2}{*}{5.1914} & 
\multirow{2}{*}{9.1775} & 
\multirow{2}{*}{0.0055} & 
\multirow{2}{*}{0.0256} & 
\multirow{2}{*}{6.3861} & 
\multirow{2}{*}{7.3846} 
\\ 

\texttt{-flash-preview-04-17} & 
& & & & & & & &
\\ \midrule

\texttt{claude-3.5} & 
\multirow{2}{*}{-0.2195} & 
\multirow{2}{*}{-12.1077} & 
\multirow{2}{*}{-7.7864} & 
\multirow{2}{*}{0.2943} &
\multirow{2}{*}{5.4107} & 
\multirow{2}{*}{8.8726} & 
\multirow{2}{*}{0.0071} & 
\multirow{2}{*}{0.0321} & 
\multirow{2}{*}{6.7834} & 
\multirow{2}{*}{7.5376} 
\\ 

\texttt{-haiku-20241022} & 
& & & & & & & &
\\ \midrule

\texttt{claude-3.7} & 
\multirow{2}{*}{-0.2154} & 
\multirow{2}{*}{-11.7673} & 
\multirow{2}{*}{-7.5305} & 
\multirow{2}{*}{0.3233} & 
\multirow{2}{*}{5.2393} & 
\multirow{2}{*}{\textbf{9.7834}} & 
\multirow{2}{*}{0.0059} & 
\multirow{2}{*}{0.0300} & 
\multirow{2}{*}{5.9620} & 
\multirow{2}{*}{7.1544} 
\\ 

\texttt{-sonnet-20250219} & 
& & & & & & & &
\\ \midrule

\texttt{qwen2.5-7b} & 
\multirow{2}{*}{-0.2049} & 
\multirow{2}{*}{-11.5481} & 
\multirow{2}{*}{-8.3208} & 
\multirow{2}{*}{0.3121} &
\multirow{2}{*}{5.7103} & 
\multirow{2}{*}{8.4121} & 
\multirow{2}{*}{0.0054} & 
\multirow{2}{*}{0.0208} & 
\multirow{2}{*}{7.3481} & 
\multirow{2}{*}{8.5285} 
\\ 

\texttt{-instruct-turbo} & 
& & & & & & & &
\\ \midrule

\texttt{qwen2.5-72b} & 
\multirow{2}{*}{-0.2066} & 
\multirow{2}{*}{-11.7391} & 
\multirow{2}{*}{-7.4165} &
\multirow{2}{*}{0.3107} & 
\multirow{2}{*}{5.3038} & 
\multirow{2}{*}{\underline{9.2297}} & 
\multirow{2}{*}{0.0056} & 
\multirow{2}{*}{0.0248} & 
\multirow{2}{*}{6.6172} & 
\multirow{2}{*}{7.8053} 
\\ 

\texttt{-instruct-turbo} & 
& & & & & & & &
\\ \midrule

\texttt{llama-3.2-3b} & 
\multirow{2}{*}{-0.1981} & 
\multirow{2}{*}{-11.6759} & 
\multirow{2}{*}{-7.2913} & 
\multirow{2}{*}{0.3152} & 
\multirow{2}{*}{5.9292} & 
\multirow{2}{*}{6.1641} & 
\multirow{2}{*}{\textbf{0.0117}} & 
\multirow{2}{*}{\textbf{0.0339}} & 
\multirow{2}{*}{8.2703} & 
\multirow{2}{*}{9.5635} 
\\ 

\texttt{-instruct-turbo} & 
& & & & & & & &
\\ \midrule

\texttt{llama-3.3-70b} & 
\multirow{2}{*}{-0.1976} & 
\multirow{2}{*}{-11.656} & 
\multirow{2}{*}{-7.3195} &
\multirow{2}{*}{0.2904} &
\multirow{2}{*}{5.2444} & 
\multirow{2}{*}{7.8648} & 
\multirow{2}{*}{0.0059} & 
\multirow{2}{*}{0.0232} & 
\multirow{2}{*}{6.6544} & 
\multirow{2}{*}{7.6213}
\\ 

\texttt{-instruct-turbo} & 
& & & & & & & &
\\ \midrule

\texttt{llama-4-maverick-17b} & 
\multirow{2}{*}{-0.2311} & 
\multirow{2}{*}{-12.313} & 
\multirow{2}{*}{-7.9958} & 
\multirow{2}{*}{0.3319} & 
\multirow{2}{*}{4.9387} & 
\multirow{2}{*}{8.8113} & 
\multirow{2}{*}{0.0059} & 
\multirow{2}{*}{0.025} & 
\multirow{2}{*}{5.5613} & 
\multirow{2}{*}{6.3629}
\\ 

\texttt{-128e-instruct-fp8} & 
& & & & & & & &
\\ \midrule

\texttt{deepseek-v3} & 
-0.1941 & 
-11.4944 & 
-7.2265 &
0.328 &
6.0934 &
8.8475 & 
0.007 & 
\underline{0.033} & 
8.6119 & 
10.1566
\\

\bottomrule[1pt]
\end{tabular*}
\label{tab:tm_unlabeled_full}
% \vspace{-3mm}
\end{table*}

\begin{table*}[t]
\centering \small
\renewcommand{\arraystretch}{0.8}
\setlength{\tabcolsep}{12pt} 
\caption{Topic modeling benchmark performance with label-based metrics. Best performance is marked as \textbf{bold}, and second-best is \underline{underlined}.}
\begin{tabular*}{\textwidth}{l!{\vrule}ccc!{\vrule}ccc} 
\toprule
 & Precision ($\uparrow$) & Recall ($\uparrow$) & F1 ($\uparrow$) & Purity ($\uparrow$) & ARI ($\uparrow$) & NMI ($\uparrow$) \\ \midrule[1pt]

\multicolumn{7}{c}{\textbf{\textit{Domain: Laptop}}} \\ \midrule[1pt]

LDA & 
0.3602 & 
0.3565 & 
0.3577 & 
0.3770 &
0.1573 & 
0.3344 
\\ \midrule

BERTopic & 
0.5139 & 
0.5084 & 
0.5102 & 
0.5033 &
0.2603 & 
0.5262 
\\ \midrule

% \texttt{4.1-nano} & 
% - & 
% - & 
% - & 
% - &
% - & 
% - 
% \\ \midrule

\texttt{gpt-4o} & 
0.6429 & 
0.6331 & 
0.6364 & 
0.6144 &
0.3380 & 
0.6095
\\ \midrule

\texttt{o3-mini} & 
0.7114 & 
0.7017 & 
0.7049 & 
0.6846 &
0.4685 & 
0.6257 
\\ \midrule

\texttt{o4-mini} & 
0.7308 & 
0.7211 & 
0.7243 & 
0.7023 &
0.4743 & 
0.6413 
\\ \midrule

\texttt{gemini-2.0} & 
\multirow{2}{*}{0.6278} & 
\multirow{2}{*}{0.6188} & 
\multirow{2}{*}{0.6218} & 
\multirow{2}{*}{0.6042} & 
\multirow{2}{*}{0.2978} & 
\multirow{2}{*}{0.6013}
\\ 

\texttt{-flash-lite} & 
& & & & & 
\\ \midrule

\texttt{gemini-2.0} & 
\multirow{2}{*}{0.7138} & 
\multirow{2}{*}{0.7036} & 
\multirow{2}{*}{0.7070} & 
\multirow{2}{*}{0.6857} & 
\multirow{2}{*}{0.4580} & 
\multirow{2}{*}{0.6315} 
\\ 

\texttt{-flash} & 
& & & & & 
\\ \midrule

\texttt{gemini-2.5} & 
\multirow{2}{*}{\underline{0.7362}} & 
\multirow{2}{*}{\underline{0.7259}} & 
\multirow{2}{*}{\underline{0.7293}} & 
\multirow{2}{*}{\underline{0.7088}} & 
\multirow{2}{*}{0.4622} & 
\multirow{2}{*}{0.6508} 
\\ 

\texttt{-flash-preview-04-17} & 
& & & & & 
\\ \midrule

\texttt{claude-3.5} & 
\multirow{2}{*}{0.5811} & 
\multirow{2}{*}{0.5722} & 
\multirow{2}{*}{0.5752} & 
\multirow{2}{*}{0.5661} & 
\multirow{2}{*}{0.2565} & 
\multirow{2}{*}{0.5845}
\\ 

\texttt{-haiku-20241022} & 
& & & & & 
\\ \midrule

\texttt{claude-3.7} & 
\multirow{2}{*}{0.7250} & 
\multirow{2}{*}{0.7148} & 
\multirow{2}{*}{0.7182} & 
\multirow{2}{*}{0.7063} & 
\multirow{2}{*}{0.4411} & 
\multirow{2}{*}{\underline{0.6530}}
\\ 

\texttt{-sonnet-20250219} & 
& & & & & 
\\ \midrule

\texttt{qwen2.5-7b} & 
\multirow{2}{*}{0.6215} & 
\multirow{2}{*}{0.6127} & 
\multirow{2}{*}{0.6156} & 
\multirow{2}{*}{0.6199} & 
\multirow{2}{*}{0.3794} & 
\multirow{2}{*}{0.5710}
\\

\texttt{-instruct-turbo} & 
& & & & & 
\\ \midrule

\texttt{qwen2.5-72b} & 
\multirow{2}{*}{0.7172} & 
\multirow{2}{*}{0.7068} & 
\multirow{2}{*}{0.7102} & 
\multirow{2}{*}{0.6921} & 
\multirow{2}{*}{0.4490} & 
\multirow{2}{*}{0.6375}
\\ 

\texttt{-instruct-turbo} & 
& & & & & 
\\ \midrule

\texttt{llama-3.2-3b} & 
\multirow{2}{*}{0.3839} & 
\multirow{2}{*}{0.3768} & 
\multirow{2}{*}{0.3792} & 
\multirow{2}{*}{0.3770} & 
\multirow{2}{*}{0.1144} & 
\multirow{2}{*}{0.3675}
\\ 

\texttt{-instruct-turbo} & 
& & & & & 
\\ \midrule

\texttt{llama-3.3-70b} & 
\multirow{2}{*}{0.7318} & 
\multirow{2}{*}{0.7213} & 
\multirow{2}{*}{0.7248} & 
\multirow{2}{*}{0.7030} & 
\multirow{2}{*}{\textbf{0.4882}} & 
\multirow{2}{*}{0.6342}
\\ 

\texttt{-instruct-turbo} & 
& & & & & 
\\ \midrule

\texttt{llama-4-maverick-17b} & 
\multirow{2}{*}{0.6997} & 
\multirow{2}{*}{0.6902} & 
\multirow{2}{*}{0.6934} & 
\multirow{2}{*}{0.6816} & 
\multirow{2}{*}{0.4479} & 
\multirow{2}{*}{0.6112}
\\ 

\texttt{-128e-instruct-fp8} & 
& & & & & 
\\ \midrule

\texttt{deepseek-v3} & 
\textbf{0.7454} & 
\textbf{0.7347} & 
\textbf{0.7383} & 
\textbf{0.7208} &
\underline{0.4814} & 
\textbf{0.6543}
\\ \midrule[1pt]

\multicolumn{7}{c}{\textbf{\textit{Domain: Restaurant}}} \\ \midrule[1pt]

LDA & 
0.4530 & 
0.4505 & 
0.4512 & 
0.4812 &
0.1994 & 
0.2397 
\\ \midrule

BERTopic & 
0.6728 & 
0.6679 & 
0.6692 & 
0.6334 &
0.4593 & 
0.5190 
\\ \midrule

% \texttt{4.1-nano} & 
% - & 
% - & 
% - & 
% - &
% - & 
% - 
% \\ \midrule

\texttt{gpt-4o} & 
0.8045 & 
0.8001 & 
0.8015 & 
0.7932 &
0.6377 & 
0.6648 
\\ \midrule

\texttt{o3-mini} & 
0.8276 & 
0.8226 & 
0.8245 & 
0.8068 &
0.6648 & 
0.6693 
\\ \midrule

\texttt{o4-mini} & 
0.8172 & 
0.8124 & 
0.8139 & 
0.7984 &
0.6384 & 
0.6640 
\\ \midrule

\texttt{gemini-2.0} & 
\multirow{2}{*}{0.8132} & 
\multirow{2}{*}{0.8077} & 
\multirow{2}{*}{0.8095} & 
\multirow{2}{*}{0.7902} & 
\multirow{2}{*}{0.6402} & 
\multirow{2}{*}{0.6706} 
\\ 

\texttt{-flash-lite} & 
& & & & & 
\\ \midrule

\texttt{gemini-2.0} & 
\multirow{2}{*}{0.8057} & 
\multirow{2}{*}{0.8007} & 
\multirow{2}{*}{0.8023} & 
\multirow{2}{*}{0.7861} & 
\multirow{2}{*}{0.6184} & 
\multirow{2}{*}{0.6609}
\\

\texttt{-flash} & 
& & & & & 
\\ \midrule

\texttt{gemini-2.5} & 
\multirow{2}{*}{0.8149} & 
\multirow{2}{*}{0.8099} & 
\multirow{2}{*}{0.8115} & 
\multirow{2}{*}{0.8019} & 
\multirow{2}{*}{0.6426} & 
\multirow{2}{*}{0.6753} 
\\ 

\texttt{-flash-preview-04-17} & 
& & & & & 
\\ \midrule

\texttt{claude-3.5} & 
\multirow{2}{*}{0.6987} & 
\multirow{2}{*}{0.6931} & 
\multirow{2}{*}{0.6948} & 
\multirow{2}{*}{0.6913} & 
\multirow{2}{*}{0.5019} & 
\multirow{2}{*}{0.5888} 
\\ 

\texttt{-haiku-20241022} & 
& & & & & 
\\ \midrule

\texttt{claude-3.7} & 
\multirow{2}{*}{\textbf{0.8386}} & 
\multirow{2}{*}{\textbf{0.8336}} & 
\multirow{2}{*}{\textbf{0.8353}} & 
\multirow{2}{*}{\textbf{0.8224}} & 
\multirow{2}{*}{\textbf{0.6805}} & 
\multirow{2}{*}{\textbf{0.6943}} 
\\ 

\texttt{-sonnet-20250219} & 
& & & & & 
\\ \midrule

\texttt{qwen2.5-7b} & 
\multirow{2}{*}{0.7264} & 
\multirow{2}{*}{0.7201} & 
\multirow{2}{*}{0.7220} & 
\multirow{2}{*}{0.7027} & 
\multirow{2}{*}{0.5212} & 
\multirow{2}{*}{0.6001}
\\ 

\texttt{-instruct-turbo} & 
& & & & & 
\\ \midrule

\texttt{qwen2.5-72b} & 
\multirow{2}{*}{0.7889} & 
\multirow{2}{*}{0.7842} & 
\multirow{2}{*}{0.7857} & 
\multirow{2}{*}{0.7806} & 
\multirow{2}{*}{0.6280} & 
\multirow{2}{*}{0.6552}
\\ 

\texttt{-instruct-turbo} & 
& & & & & 
\\ \midrule

\texttt{llama-3.2-3b} & 
\multirow{2}{*}{0.4789} & 
\multirow{2}{*}{0.4743} & 
\multirow{2}{*}{0.4756} & 
\multirow{2}{*}{0.5380} & 
\multirow{2}{*}{0.3277} & 
\multirow{2}{*}{0.3955}
\\ 

\texttt{-instruct-turbo} & 
& & & & & 
\\ \midrule

\texttt{llama-3.3-70b} & 
\multirow{2}{*}{0.7912} & 
\multirow{2}{*}{0.7858} & 
\multirow{2}{*}{0.7875} & 
\multirow{2}{*}{0.7667} & 
\multirow{2}{*}{0.6207} & 
\multirow{2}{*}{0.6795}
\\ 

\texttt{-instruct-turbo} & 
& & & & & 
\\ \midrule

\texttt{llama-4-maverick-17b} & 
\multirow{2}{*}{0.8109} & 
\multirow{2}{*}{0.8061} & 
\multirow{2}{*}{0.8077} & 
\multirow{2}{*}{0.7961} & 
\multirow{2}{*}{0.6525} & 
\multirow{2}{*}{0.6644}
\\ 

\texttt{-128e-instruct-fp8} & 
& & & & & 
\\ \midrule

\texttt{deepseek-v3} & 
\underline{0.8317} & 
\underline{0.8260} & 
\underline{0.8278} & 
\underline{0.8113} &
\underline{0.6737} & 
\underline{0.6938} 
\\

\bottomrule[1pt]
\end{tabular*}
\label{tab:tm_labeled_full}
% \vspace{-3mm}
\end{table*}

%% file: tables/inter_annotator.tex
\begin{table}[ht]
\centering
\caption{Cohen's Kappa scores for inter-annotator agreement.}
\begin{tabular}{lcccc}
\toprule
\textbf{Domain} & \textbf{easy\_single} & \textbf{easy\_double} & \textbf{hard\_single} & \textbf{hard\_double} \\
\midrule
\textbf{Laptop} & 1.0000 & 1.0000 & 0.9519 & 0.8650 \\
\textbf{Restaurant} & 0.9514 & 0.9550 & 0.9753 & 0.8800 \\
\bottomrule
\end{tabular}
\label{tab:iaa-kappa}
\end{table}

%% file: tables/qit_appendix.tex
\subsection{F1}

\begin{table*}[ht]
\centering \small
\renewcommand{\arraystretch}{0.8}
\setlength{\tabcolsep}{13pt} 
\caption{
Benchmark results of segment intrusion tasks with F1 metric. Best performance is marked as \textbf{bold}, and second-best is \underline{underlined}.
}
\begin{tabular*}{\textwidth}{c!{\vrule}ccccc} 
\toprule
 & \textbf{SI-E} & \textbf{SI-H} & \textbf{DI-E} & \textbf{DI-H} & \textbf{Avg. F1} \\ \midrule[1pt]

\multicolumn{6}{c}{\textbf{\textit{Domain: Laptop}}} \\ \midrule[1pt]

\texttt{Human Performance} & 
1.0000 & 
0.9700 & 
0.9900 &
0.8700 &
0.9575
\\ \midrule

\texttt{gpt-4o} & 
0.9700 & 
0.8550 & 
0.9500 &
0.7734 &
0.8871
\\ \midrule

\texttt{o3-mini} & 
0.9800 & 
\textbf{0.9050} & 
0.9450 &
\underline{0.8099} &
\underline{0.9100}
\\ \midrule

\texttt{o4-mini} & 
\textbf{0.9900} & 
0.8800 & 
\textbf{0.9750} &
\textbf{0.8218} &
\textbf{0.9167}
\\ \midrule

\texttt{gemini-2.0-flash-lite} & 
0.8664 & 
0.7087 & 
0.9055 &
0.6793 &
0.7900
\\ \midrule

\texttt{gemini-2.0-flash} & 
0.9350 & 
0.7800 & 
0.9273 &
0.6828 &
0.8313
\\ \midrule

\texttt{gemini-2.5-flash-preview-04-17} & 
\underline{0.9850} & 
\underline{0.8950} & 
0.8179 &
0.6648 &
0.8407
\\ \midrule

\texttt{claude-3.5-haiku-20241022} & 
0.9167 & 
0.7922 & 
0.8878 &
0.6469 &
0.8109
\\ \midrule

\texttt{claude-3.7-sonnet-20250219} & 
0.9650 & 
0.8250 & 
0.9450 &
0.8089 &
0.8860
\\ \midrule

\texttt{qwen2.5-7b-instruct-turbo} & 
0.8515 & 
0.6350 & 
0.6000 &
0.3014 &
0.5970
\\ \midrule

\texttt{qwen2.5-72b-instruct-turbo} & 
0.9676 & 
0.8300 & 
\underline{0.9600} &
0.6914 &
0.8622
\\ \midrule

\texttt{llama-3.2-3b-instruct-turbo} & 
0.2625 & 
0.2510 & 
0.0921 &
0.0787 &
0.1711
\\ \midrule

\texttt{llama-3.3-70b-instruct-turbo} & 
0.9600 & 
0.8100 & 
0.9400 &
0.7111 &
0.8553
\\ \midrule

\texttt{llama-4-maverick-17b-128e-instruct-fp8} & 
0.5459 & 
0.4450 & 
0.3098 &
0.2129 &
0.3784
\\ \midrule

\texttt{deepseek-v3} & 
0.9550 & 
0.8350 & 
0.9100 &
0.7174 &
0.8544
\\ 
\midrule[1pt]

\multicolumn{6}{c}{\textbf{\textit{Domain: Restaurant}}} \\ \midrule[1pt]

\texttt{Human Performance} & 
0.9800 & 
0.9700 & 
0.9700 &
0.8700 &
0.9475
\\ \midrule

\texttt{gpt-4o} & 
0.9350 & 
0.7931 & 
0.8928 &
0.7348 &
0.8389
\\ \midrule

\texttt{o3-mini} & 
\underline{0.9750} & 
0.8212 & 
\underline{0.9400} &
\underline{0.7463} &
\underline{0.8706}
\\ \midrule

\texttt{o4-mini} & 
\textbf{0.9800} & 
\textbf{0.8371} & 
\textbf{0.9500} &
\textbf{0.7718} &
\textbf{0.8847}
\\ \midrule

\texttt{gemini-2.0-flash-lite} & 
0.7982 & 
0.6257 & 
0.7624 &
0.6667 &
0.7132
\\ \midrule

\texttt{gemini-2.0-flash} & 
0.9000 & 
0.6650 & 
0.8628 &
0.6377 &
0.7664
\\ \midrule

\texttt{gemini-2.5-flash-preview-04-17} & 
\underline{0.9750} & 
\underline{0.8250} & 
0.7627 &
0.6398 &
0.8006
\\ \midrule

\texttt{claude-3.5-haiku-20241022} & 
0.9051 & 
0.7255 & 
0.8200 &
0.6634 &
0.7785
\\ \midrule

\texttt{claude-3.7-sonnet-20250219} & 
0.9676 & 
0.7750 & 
\textbf{0.9500} &
0.7299 &
0.8556
\\ \midrule

\texttt{qwen2.5-7b-instruct-turbo} & 
0.7481 & 
0.5350 & 
0.4010 &
0.3175 &
0.5004
\\ \midrule

\texttt{qwen2.5-72b-instruct-turbo} & 
0.9127 & 
0.7400 & 
0.9104 &
0.6586 &
0.8054
\\ \midrule

\texttt{llama-3.2-3b-instruct-turbo} & 
0.2707 & 
0.2076 & 
0.1049 &
0.0750 &
0.1645
\\ \midrule

\texttt{llama-3.3-70b-instruct-turbo} & 
0.9552 & 
0.7419 & 
0.8878 &
0.6489 &
0.8084
\\ \midrule

\texttt{llama-4-maverick-17b-128e-instruct-fp8} & 
0.5594 & 
0.4559 & 
0.2148 &
0.1982 &
0.3571
\\ \midrule

\texttt{deepseek-v3} & 
0.9500 & 
0.7850 & 
0.8557 &
0.7022 &
0.8232
\\

\bottomrule[1pt]
\end{tabular*}
\label{tab:qit}
% \vspace{-3mm}
\end{table*}

\clearpage

\subsection{Recall}

\begin{table*}[ht]
\centering \small
\renewcommand{\arraystretch}{0.8}
\setlength{\tabcolsep}{13pt} 
\caption{
Benchmark results of segment intrusion tasks with Recall metric. Best performance is marked as \textbf{bold}, and second-best is \underline{underlined}.
}
\begin{tabular*}{\textwidth}{c!{\vrule}ccccc} 
\toprule
 & \textbf{SI-E} & \textbf{SI-H} & \textbf{DI-E} & \textbf{DI-H} & \textbf{Avg. R} \\ \midrule[1pt]

\multicolumn{6}{c}{\textbf{\textit{Domain: Laptop}}} \\ \midrule[1pt]

\texttt{Human Performance} & 
1.0000 & 
0.9700 & 
0.9900 &
0.8700 &
0.9575
\\ \midrule

\texttt{gpt-4o} & 
0.9700 & 
0.8550 & 
0.9500 & 
0.7850 & 
0.8900
\\ \midrule

\texttt{o3-mini} & 
0.9800 & 
\textbf{0.9050} & 
0.9450 & 
\underline{0.8200} & 
\underline{0.9125}
\\ \midrule

\texttt{o4-mini} & 
\textbf{0.9900} & 
0.8800 & 
\textbf{0.9750} & 
\textbf{0.8300} & 
\textbf{0.9188}
\\ \midrule

\texttt{gemini-2.0-flash-lite} & 
0.9400 & 
0.8150 & 
0.9100 & 
0.7150 & 
0.8450
\\ \midrule

\texttt{gemini-2.0-flash} & 
0.9350 & 
0.7800 & 
0.9250 & 
0.7050 & 
0.8363
\\ \midrule

\texttt{gemini-2.5-flash-preview-04-17} & 
\underline{0.9850} & 
\underline{0.8950} & 
0.7300 & 
0.6000 & 
0.8025
\\ \midrule

\texttt{claude-3.5-haiku-20241022} & 
0.9350 & 
0.8100 & 
0.8900 & 
0.6550 & 
0.8225
\\ \midrule

\texttt{claude-3.7-sonnet-20250219} & 
0.9650 & 
0.8250 & 
0.9450 & 
0.8150 & 
0.8875
\\ \midrule

\texttt{qwen2.5-7b-instruct-turbo} & 
0.8600 & 
0.6350 & 
0.6000 & 
0.3150 & 
0.6025
\\ \midrule

\texttt{qwen2.5-72b-instruct-turbo} & 
0.9700 & 
0.8300 & 
\underline{0.9600} & 
0.7000 & 
0.8650
\\ \midrule

\texttt{llama-3.2-3b-instruct-turbo} & 
0.5000 & 
0.4950 & 
0.1250 & 
0.1100 & 
0.3075
\\ \midrule
\texttt{llama-3.3-70b-instruct-turbo} & 
0.9600 & 
0.8100 & 
0.9400 & 
0.7200 & 
0.8575
\\ \midrule

\texttt{llama-4-maverick-17b-128e-instruct-fp8} & 
0.5500 & 
0.4550 & 
0.3400 & 
0.2400 & 
0.3963
\\ \midrule

\texttt{deepseek-v3} & 
0.9550 & 
0.8350 & 
0.9100 & 
0.7300 & 
0.8575
\\ \midrule[1pt]

\multicolumn{6}{c}{\textbf{\textit{Domain: Restaurant}}} \\ \midrule[1pt]

\texttt{Human Performance} & 
0.9800 & 
0.9700 & 
0.9700 &
0.8700 &
0.9475
\\ \midrule

\texttt{gpt-4o} & 
0.9350 & 
0.8050 & 
0.8950 & 
0.7550 & 
0.8475
\\ \midrule

\texttt{o3-mini} & 
\underline{0.9750} & 
0.8150 & 
\underline{0.9400} & 
\underline{0.7650} & 
\underline{0.8738}
\\ \midrule

\texttt{o4-mini} & 
\textbf{0.9800} & 
\textbf{0.8350} & 
\textbf{0.9500} & 
\textbf{0.7950} & 
\textbf{0.8900}
\\ \midrule

\texttt{gemini-2.0-flash-lite} & 
0.8900 & 
0.7900 & 
0.7700 & 
0.7100 & 
0.7900
\\ \midrule

\texttt{gemini-2.0-flash} & 
0.9000 & 
0.6650 & 
0.8650 & 
0.6600 & 
0.7725
\\ \midrule

\texttt{gemini-2.5-flash-preview-04-17} & 
\underline{0.9750} & 
\underline{0.8250} & 
0.6750 & 
0.5950 & 
0.7675
\\ \midrule

\texttt{claude-3.5-haiku-20241022} & 
0.9300 & 
0.7400 & 
0.8200 & 
0.6800 & 
0.7925
\\ \midrule

\texttt{claude-3.7-sonnet-20250219} & 
0.9700 & 
0.7750 & 
\textbf{0.9500} & 
0.7500 & 
0.8613
\\ \midrule

\texttt{qwen2.5-7b-instruct-turbo} & 
0.7500 & 
0.5350 & 
0.4050 & 
0.3350 & 
0.5063
\\ \midrule

\texttt{qwen2.5-72b-instruct-turbo} & 
0.9150 & 
0.7400 & 
0.9150 & 
0.6800 & 
0.8125
\\ \midrule

\texttt{llama-3.2-3b-instruct-turbo} & 
0.4500 & 
0.3950 & 
0.1350 & 
0.0950 & 
0.2688
\\ \midrule

\texttt{llama-3.3-70b-instruct-turbo} & 
0.9600 & 
0.7400 & 
0.8900 & 
0.6700 & 
0.8150
\\ \midrule

\texttt{llama-4-maverick-17b-128e-instruct-fp8} & 
0.5650 & 
0.4650 & 
0.2400 & 
0.2250 & 
0.3738
\\ \midrule

\texttt{deepseek-v3} & 
0.9500 & 
0.7850 & 
0.8600 & 
0.7250 & 
0.8300
\\

\bottomrule[1pt]
\end{tabular*}
\label{tab:qit_r}
% \vspace{-3mm}
\end{table*}

\clearpage
\subsection{Precision}

\begin{table*}[ht]
\centering \small
\renewcommand{\arraystretch}{0.8}
\setlength{\tabcolsep}{13pt} 
\caption{
Benchmark results of segment intrusion tasks with Precision metric. Best performance is marked as \textbf{bold}, and second-best is \underline{underlined}.
}
\begin{tabular*}{\textwidth}{c!{\vrule}ccccc} 
\toprule
 & \textbf{SI-E} & \textbf{SI-H} & \textbf{DI-E} & \textbf{DI-H} & \textbf{Avg. P} \\ \midrule[1pt]

\multicolumn{6}{c}{\textbf{\textit{Domain: Laptop}}} \\ \midrule[1pt]

\texttt{Human Performance} & 
1.0000 & 
0.9700 & 
0.9900 &
0.8700 &
0.9575
\\ \midrule

\texttt{gpt-4o} & 
0.9700 & 
0.8550 & 
0.9500 & 
0.7621 & 
0.8843
\\ \midrule

\texttt{o3-mini} & 
0.9800 & 
\textbf{0.9050} & 
0.9450 & 
0.8000 & 
\underline{0.9075}
\\ \midrule

\texttt{o4-mini} & 
\textbf{0.9900} & 
0.8800 & 
\textbf{0.9750} & 
\textbf{0.8137} & 
\textbf{0.9147}
\\ \midrule

\texttt{gemini-2.0-flash-lite} & 
0.8034 & 
0.6269 & 
0.9010 & 
0.6471 & 
0.7446
\\ \midrule

\texttt{gemini-2.0-flash} & 
0.9350 & 
0.7800 & 
0.9296 & 
0.6620 & 
0.8267
\\ \midrule

\texttt{gemini-2.5-flash-preview-04-17} & 
\underline{0.9850} & 
\underline{0.8950} & 
0.9299 & 
0.7453 & 
0.8888
\\ \midrule

\texttt{claude-3.5-haiku-20241022} & 
0.8990 & 
0.7751 & 
0.8856 & 
0.6390 & 
0.7997
\\ \midrule

\texttt{claude-3.7-sonnet-20250219} & 
0.9650 & 
0.8250 & 
0.9450 & 
\underline{0.8030} & 
0.8845
\\ \midrule

\texttt{qwen2.5-7b-instruct-turbo} & 
0.8431 & 
0.6350 & 
0.6000 & 
0.2890 & 
0.5918
\\ \midrule

\texttt{qwen2.5-72b-instruct-turbo} & 
0.9652 & 
0.8300 & 
\underline{0.9600} & 
0.6829 & 
0.8595
\\ \midrule

\texttt{llama-3.2-3b-instruct-turbo} & 
0.1779 & 
0.1681 & 
0.0729 & 
0.0613 & 
0.1201
\\ \midrule

\texttt{llama-3.3-70b-instruct-turbo} & 
0.9600 & 
0.8100 & 
0.9400 & 
0.7024 & 
0.8531
\\ \midrule

\texttt{llama-4-maverick-17b-128e-instruct-fp8} & 
0.5419 & 
0.4354 & 
0.2845 & 
0.1912 & 
0.3633
\\ \midrule

\texttt{deepseek-v3} & 
0.9550 & 
0.8350 & 
0.9100 & 
0.7053 & 
0.8513
\\ \midrule[1pt]

\multicolumn{6}{c}{\textbf{\textit{Domain: Restaurant}}} \\ \midrule[1pt]

\texttt{Human Performance} & 
0.9800 & 
0.9700 & 
0.9700 &
0.8700 &
0.9475
\\ \midrule

\texttt{gpt-4o} & 
0.9350 & 
0.7816 & 
0.8905 & 
0.7156 & 
0.8307
\\ \midrule

\texttt{o3-mini} & 
\underline{0.9750} & 
\underline{0.8274} & 
\underline{0.9400} & 
0.7286 & 
\underline{0.8678}
\\ \midrule

\texttt{o4-mini} & 
\textbf{0.9800} & 
\textbf{0.8392} & 
\textbf{0.9500} & 
\underline{0.7500} & 
\textbf{0.8798}
\\ \midrule

\texttt{gemini-2.0-flash-lite} & 
0.7236 & 
0.5180 & 
0.7549 & 
0.6283 & 
0.6562
\\ \midrule

\texttt{gemini-2.0-flash} & 
0.9000 & 
0.6650 & 
0.8607 & 
0.6168 & 
0.7606
\\ \midrule

\texttt{gemini-2.5-flash-preview-04-17} & 
\underline{0.9750} & 
0.8250 & 
0.8766 & 
0.6919 & 
0.8421
\\ \midrule

\texttt{claude-3.5-haiku-20241022} & 
0.8815 & 
0.7115 & 
0.8200 & 
0.6476 & 
0.7652
\\ \midrule

\texttt{claude-3.7-sonnet-20250219} & 
0.9652 & 
0.7750 & 
\textbf{0.9500} & 
\textbf{0.7709} & 
0.8653
\\ \midrule

\texttt{qwen2.5-7b-instruct-turbo} & 
0.7463 & 
0.5350 & 
0.3971 & 
0.3018 & 
0.4951
\\ \midrule

\texttt{qwen2.5-72b-instruct-turbo} & 
0.9104 & 
0.7400 & 
0.9059 & 
0.6385 & 
0.7987
\\ \midrule

\texttt{llama-3.2-3b-instruct-turbo} & 
0.1935 & 
0.1408 & 
0.0857 & 
0.0619 & 
0.1205
\\ \midrule

\texttt{llama-3.3-70b-instruct-turbo} & 
0.9505 & 
0.7437 & 
0.8856 & 
0.6291 & 
0.8022
\\ \midrule

\texttt{llama-4-maverick-17b-128e-instruct-fp8} & 
0.5539 & 
0.4471 & 
0.1943 & 
0.1772 & 
0.3431
\\ \midrule

\texttt{deepseek-v3} & 
0.9500 & 
0.7850 & 
0.8515 & 
0.6808 & 
0.8168
\\

\bottomrule[1pt]
\end{tabular*}
\label{tab:qit_p}
% \vspace{-3mm}
\end{table*}

%% file: figures/qit_full_appendix.tex
\begin{figure}[htbp!]
    \centering
    \includegraphics[width=\columnwidth]{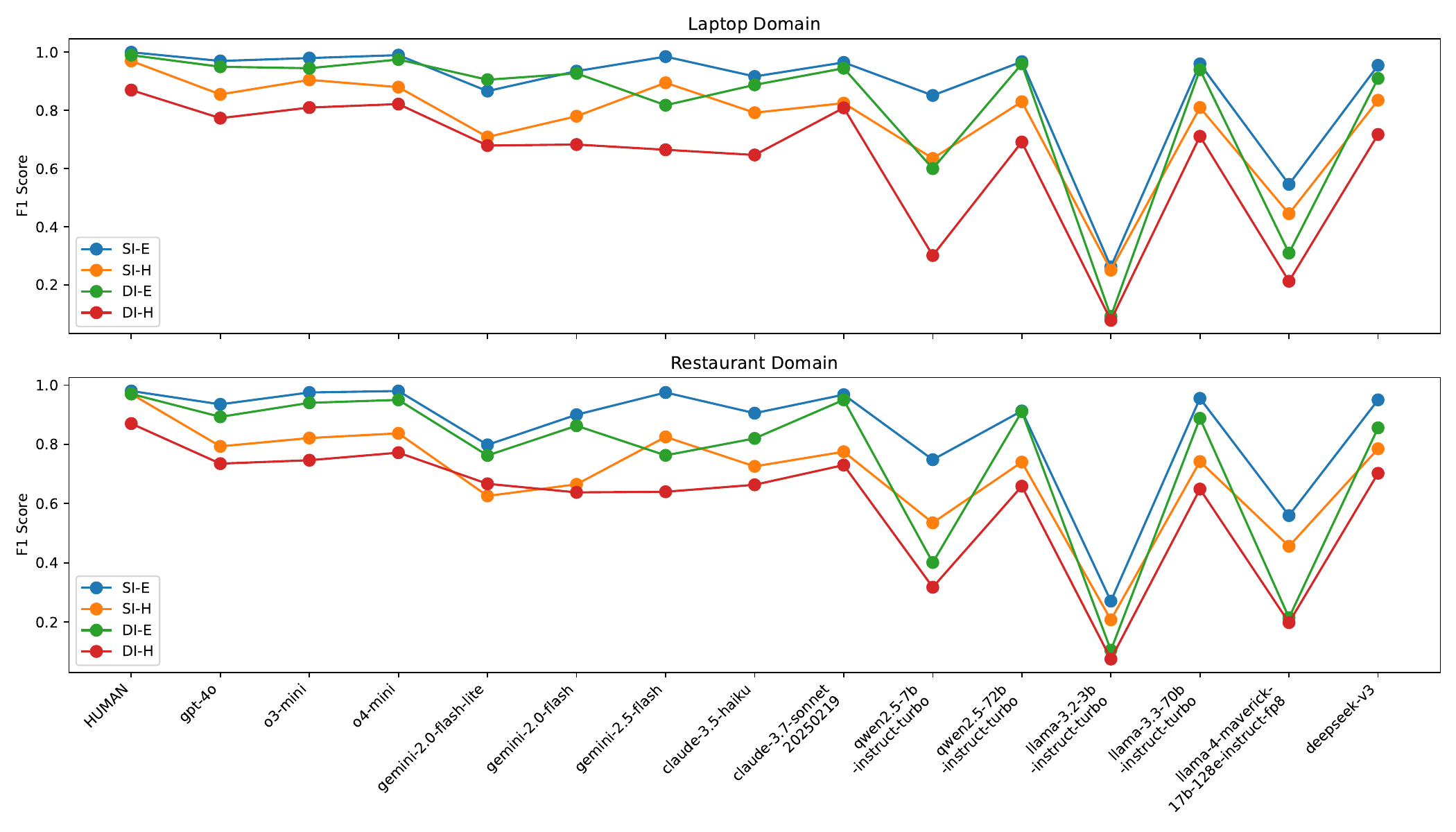}
    \caption{
        Visualized F1 performance of segment intrusion tasks. Human performance is calculated by averaging the annotations of two participants. 50 out of 200 instances for each task are randomly sampled in human evaluation.
    }
    \label{fig:qitfull}
    % \vspace{-3.7mm}
\end{figure}